\documentclass[conference]{IEEEtran}
\IEEEoverridecommandlockouts
\usepackage{cite}
\usepackage{amsmath,amssymb,amsfonts}
\usepackage{algorithmic}
\usepackage{graphicx}
\usepackage{textcomp}
\usepackage{xcolor}
\usepackage{url}
\usepackage{booktabs,multirow,colortbl}
\usepackage{listings}
\usepackage{placeins}
\usepackage{stfloats}
\usepackage{balance}
\usepackage[normalem]{ulem}
\usepackage[hidelinks]{hyperref}

\newcommand{\citep}[1]{\cite{#1}}
\newcommand{\method}{\textup{ERDR}}
\newcommand{\train}{\mathrm{train}}
\newcommand{\val}{\mathrm{val}}

\definecolor{ilsdelete}{RGB}{190,90,0}
\definecolor{ilsinsert}{RGB}{0,115,65}

\newcommand{\mstd}[2]{{\footnotesize #1}{\scriptsize$\pm$#2}}

\definecolor{taregood}{RGB}{0,96,160}
\definecolor{tarebad}{RGB}{170,70,55}
\definecolor{tareestimate}{RGB}{126,78,148}
\definecolor{tarelight}{RGB}{244,247,249}

\newcommand{\ablationratio}[2]{#1\,{\scriptsize(#2$\times$)}}
\title{Accelerating Dynamic Route Re-solving via Robust Solver Evolution in Automatic Heuristic Design}
\author{\IEEEauthorblockN{Yizhuo Ma\textsuperscript{1}, Haoran Sun\textsuperscript{1}, Shuang Liang\textsuperscript{1}, Muquan Li\textsuperscript{1},\\
Qizhi Chen\textsuperscript{1},Renzhong Wang\textsuperscript{1}, Nan Chen\textsuperscript{2}, Lunxu Liu\textsuperscript{2}, Ke Qin\textsuperscript{1*}}
\IEEEauthorblockA{\textsuperscript{1}University of Electronic Science and Technology of China, Chengdu, China\\
\textsuperscript{2}Department of Thoracic Surgery, West China Hospital, Sichuan University\\
\{myz, 2023091203011, muquanli2023, chenqizhi, 202611090902\}@std.uestc.edu.cn;\\
\{shuangliang, qinke\}@uestc.edu.cn; \{dr.chennan, liulunxu\}@wchscu.cn}
\thanks{\makebox[0.7em][l]{\textsuperscript{*}}Ke Qin is the corresponding author.\newline
\hspace*{1em}\protect\hypertarget{code-availability}{\makebox[0.7em][l]{\textsuperscript{\ensuremath{\dagger}}}The code is available at \url{https://github.com/myz12138/ERDR}.}}}

\begin{document}

% Consistent positive spacing around single- and double-column floats.
\setlength{\floatsep}{4pt plus 1pt minus 1pt}
\setlength{\dblfloatsep}{4pt plus 1pt minus 1pt}
\setlength{\textfloatsep}{6pt plus 1pt minus 1pt}
\setlength{\dbltextfloatsep}{6pt plus 1pt minus 1pt}
\setlength{\intextsep}{6pt plus 1pt minus 1pt}
\setcounter{topnumber}{4}
\setcounter{bottomnumber}{3}
\setcounter{totalnumber}{6}
\setcounter{dbltopnumber}{3}
\renewcommand{\topfraction}{0.95}
\renewcommand{\bottomfraction}{0.95}
\renewcommand{\dbltopfraction}{0.95}
\renewcommand{\textfraction}{0.05}
\renewcommand{\floatpagefraction}{0.85}
\renewcommand{\dblfloatpagefraction}{0.85}
\flushbottom

% BEGIN PRESERVED MAIN CONTENT

\maketitle

\begin{abstract}
\looseness=-1 Multi-goal route planning uses spatial data on service locations and pairwise travel costs to determine a visiting order that minimizes total travel cost. This task can be formulated as the traveling salesman problem (TSP) and has benefited from large language model-based automatic heuristic design (AHD) for constructing effective solvers.  Despite their effectiveness in static settings, existing methods exhibit substantial re-solving inefficiency when location changes, new requests, or request cancellations alter the underlying request data during execution. Our empirical analysis suggests that ``static-overfitting'' contributes to this inefficiency, as evolved solvers may repeatedly direct search toward overlapping local neighborhoods after dynamic changes. We further find that this ``overfitting'' is associated with limitations in existing evolutionary processes, where lower objective values on some instances coincide with longer solving times. In this paper, we therefore propose ERDR, a plug-and-play framework that integrates two strategies into existing AHD pipelines to guide evolution toward solvers that re-solve dynamically changed instances faster. Specifically, 1) ERDR uses validation-guided reselection to reevaluate candidates on instances excluded from routine evolution, reducing specialization to fixed training instances. 2) ERDR applies runtime robustness-augmented evaluation, supplementing the original evaluation objective of AHD methods with inter-group variability and tail risk in solving time to discourage costly and uneven search. Experiments across multiple TSP scales and dynamic scenarios show that ERDR achieves average re-solving speedups of $2.2\times$ under guided local search and $4.4\times$ under iterated local search.\hyperlink{code-availability}{\textsuperscript{\ensuremath{\dagger}}}
\end{abstract}

\begin{figure}[!htb]
  \centering
  \includegraphics[width=\columnwidth,pagebox=cropbox]{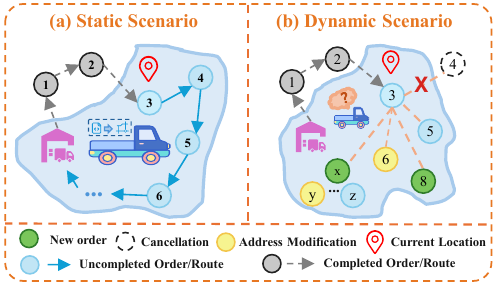}
  \par
  \includegraphics[width=\columnwidth]{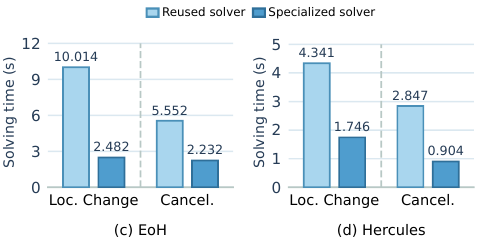}
  \caption{(a) Static execution, where an AHD-evolved solver plans a tour that is followed without instance changes. (b) Dynamic execution, where location changes, new requests, or request cancellations alter the remaining routing problem and require reuse of the evolved solver for re-solving. (c) and (d) compare mean tour re-solving times within GLS for reused and specialized solvers generated by EoH and Hercules, respectively, under location changes and request cancellations at 20\% tour progress on 50 TSP500 instances.}
  \label{fig:framework}
  \label{fig:observation}
\end{figure}

% Uncomment the following to link to your code, datasets, an extended version or similar.
% You must keep this block between (not within) the abstract and the main body of the paper.
% Make sure that you do not de-anonymize yourself with these links.
% \begin{links}
%     \link{Code}{https://aaai.org/example/code}
%     \link{Datasets}{https://aaai.org/example/datasets}
%     \link{Extended version}{https://aaai.org/example/extended-version}
% \end{links}

\section{Introduction}

\looseness=-1 Multi-goal route planning is a core computational task in location-based services such as delivery and field operations. It uses spatial data describing target locations and travel costs to determine a visiting order that minimizes travel cost, a task that can be formulated as the traveling salesman problem (TSP)~\citep{dantzig1954solution,laporte1992traveling}. Given the NP-hardness of TSP~\citep{karp1972reducibility}, solving such problems within practical computational budgets often relies on heuristic solvers to obtain high-quality routes. However, developing effective heuristics and configuring them for different problem settings requires substantial expertise and effort~\citep{hutter2009paramils}. Built on Transformer architectures~\citep{vaswani2017attention}, modern large language models (LLMs) exhibit broad few-shot capabilities~\citep{brown2020language} and improved instruction following~\citep{ouyang2022training}, including program synthesis from natural-language specifications~\citep{chen2021codex,li2022alphacode}. These advances have enabled automatic heuristic design (AHD) methods~\citep{liu2024eoh,ye2024reevo} that discover effective solvers through iterative generation, revision, and evaluation within an evolutionary framework.

\looseness=-1 Existing LLM-based AHD methods typically evolve heuristics on static data with fixed node sets and locations. As illustrated in Figure~\ref{fig:framework}~(a), the evolved solver plans a tour for the known service nodes, which the vehicle then follows to serve them in sequence. During route execution, changes in service requests update the underlying spatial data through changes to unvisited service locations, new requests, and request cancellations~\citep{horvath2025framework}. As shown in Figure~\ref{fig:framework}~(b), the remaining route must then be reoptimized from the vehicle's current location using the updated information. Such route updates are central to dynamic routing and widespread in real-world operations, including courier delivery and field maintenance~\citep{pillac2013review}. For example, tabu search~\citep{gendreau1999parallel} reoptimizes courier routes as new customer requests arrive while vehicles are in transit. RpALNS~\citep{pillac2018fast} uses parallel adaptive large neighborhood search to update technician routes as new service requests become known. However, our investigation with guided local search (GLS, see details in Section~\ref{sec:search-frameworks}) shows that directly reused solvers require substantially more solving time than counterparts evolved on the changed data. This additional solving cost delays route updates in response to changing request data. As shown in Figure~\ref{fig:observation}~(c)--(d), under location changes and request cancellations at 20\% tour progress on TSP500, directly re-solving the tour with the EoH/Hercules-evolved solvers takes, on average, 3.3/2.8 times as long as with their specialized counterparts.

\looseness=-1 Our analysis in Section~\ref{sec:observation-overfitted-search} suggests that static overfitting contributes to the re-solving inefficiency observed above, as evolution favors heuristics tailored to the structures of static source instances that may struggle to maintain stable search behavior after instance changes. During re-solving, the reused solver repeatedly directs the search toward overlapping local neighborhoods, incurring additional search cost to obtain high-quality solutions. To understand how this ``overfitting'' emerges during evolution, we identify two critical findings: 1) improvements in solution quality increasingly concentrate on a few source instances, allowing aggregate quality to improve while most instances receive little benefit. 2) these gains arise mainly on instances with longer solving times, indicating that quality improvement is coupled with increasingly costly and uneven search. These findings suggest that existing evolutionary processes tend to retain solvers whose gains are confined to a narrow subset of instances and accompanied by increasingly costly search, contributing to inefficient re-solving. We will elaborate on specific details in Section~\ref{sec:observation-source-imbalance}.

\looseness=-1 We thus introduce ERDR (\underline{E}fficient \underline{R}e-solving in \underline{D}ynamic Scenarios via \underline{R}obust Solver Evolution), a plug-and-play framework that integrates with existing AHD methods. Considering that validation sets are commonly used in machine learning to mitigate ``overfitting'' to training data, ERDR firstly introduces validation-guided reselection to reduce candidate specialization to particular instances. Under this mechanism, the instances are partitioned into disjoint training and validation groups, with routine evolution conducted on the training groups and candidates reevaluated on the validation groups to update the population. ERDR further introduces runtime robustness-augmented evaluation, which supplements original evaluation objective of AHD methods with inter-group variability and tail risk in solving-time. These criteria incorporate the associated computational costs into evolutionary selection, encouraging stable and efficient search across instances. These two strategies form a robust evaluation and selection mechanism designed to improve solver transferability and search stability when guiding tour re-solving in the dynamic scenario.

Our contributions are summarized as follows:
\begin{enumerate}

\item We identify substantial re-solving overhead when re-solving tours in dynamic TSP scenarios: solvers evolved under static conditions require substantially longer solving times when reused after instance changes during tour execution.

\item Our analysis suggests that ``static-overfitting'' contributes to this failure and shows that mean-based evaluation allows solvers with quality gains concentrated on a narrow subset of instances and increasingly costly search to remain competitive during evolution.

\item We propose ERDR, a plug-and-play evaluation framework that combines validation-guided reselection to reduce specialization to fixed training instances with runtime robustness-augmented evaluation to accelerate dynamic re-solving.

\item Experiments across multiple dynamic TSP scenarios show that ERDR can be integrated into different AHD methods, achieving average tour re-solving speedups of $2.2\times$ under GLS and $4.4\times$ under ILS.
\end{enumerate}

\section{Related work}

\paragraph{Automatic heuristic design}
\looseness=-1 Classical TSP heuristics such as guided local search (GLS)~\citep{voudouris1999gls} and Lin--Kernighan search~\citep{helsgaun2000lkh} rely on expert-designed~\citep{ma2026topollm} neighborhoods and update rules~\citep{li2025beyondrandom}. Hyper-heuristics~\citep{burke2013hyperheuristics} search over heuristics, while FunSearch~\citep{romeraparedes2024funsearch} combines LLM-generated code~\citep{wang2025graphtool} with systematic evaluation~\citep{li2026asag}. Recent AHD methods adopt different evolutionary mechanisms~\citep{wang2026dash}. EoH~\citep{liu2024eoh} co-evolves ideas and code, ReEvo~\citep{ye2024reevo} uses reflective feedback, Hercules~\citep{wu2025hercules} derives search directions from elite heuristics, and MEoH~\citep{yao2025meoh} applies multi-objective selection. Other studies broaden exploration through MCTS, joint algorithm-model evolution, and hindsight and foresight prompting~\citep{zheng2025mcts,huang2026calm,chen2026hifo}. These methods evaluate candidates on fixed instances and do not examine solver reuse after instance changes.

\paragraph{Dynamic Scenarios in TSP}
 Dynamic routing methods update routes as new information becomes available during execution~\citep{pillac2013review}. Existing routes can be adapted to updated requests and used as starting solutions for further optimization. TSP reoptimization has been studied under node insertion and removal, using an optimal tour of the original instance~\citep{archetti2003reoptimizing}. Insertion-based route planning incorporates newly arriving requests into existing routes~\citep{tong2018unified}. Other approaches retain information across changing environments. Explicit memory stores previous solutions for reuse~\citep{tinos2014memory}, while ACO-based methods combine search information with local improvement~\citep{mavrovouniotis2017dynamic}. A scheme library-based ACO method maintains precomputed route schemes for recurring or similar node-location configurations and combines their reuse with 2-opt local search~\citep{wang2023scheme}. These studies adapt routes through manually designed heuristics and reuse mechanisms. In comparison, our work focuses on enabling heuristics automatically designed by LLMs to be efficiently reused when locations or requests change during route execution.

\begin{figure*}[!tbp]
\centering
\resizebox{\textwidth}{!}{%
\includegraphics[height=0.201\textwidth]{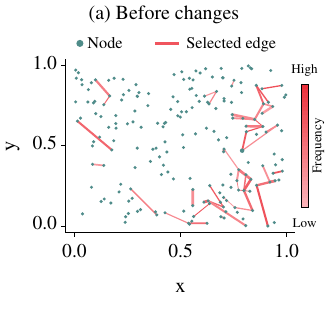}\hspace{0.005\textwidth}%
\includegraphics[height=0.201\textwidth]{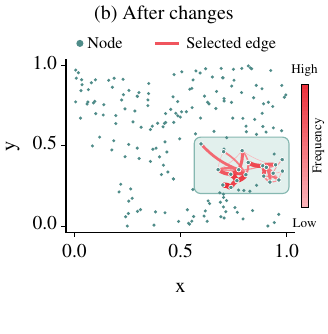}\hspace{0.015\textwidth}%
\includegraphics[height=0.201\textwidth]{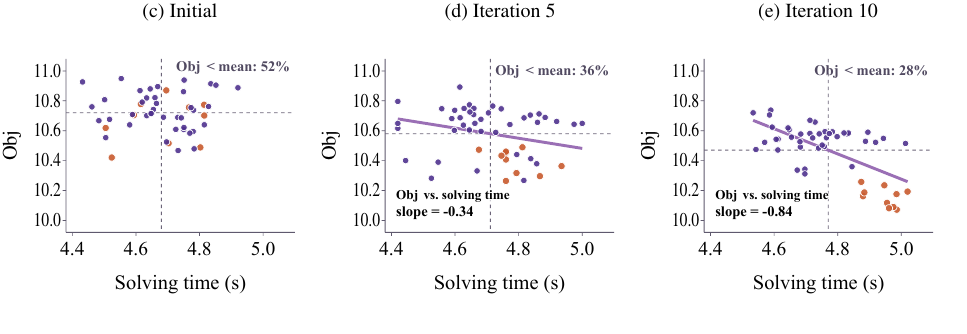}
}

\caption{Experimental observations of re-solving inefficiency within GLS. (a)--(b) show edge selections before and after dynamic changes, respectively, recorded over at most 500 GLS iterations using the EoH solver evolved on static TSP200. (a) shows the original 200-node instance, while (b) shows the remaining task after 20\% tour progress, with new nodes added at 20\% of the unvisited-node count. Points denote TSP nodes, and edge darkness and width encode selection frequency during solving. (c)--(e) show instance-level joint distributions of objective value and solving time over 50 source TSP200 instances at initialization and evolution iterations 5 and 10. Each point denotes one instance. In (d) and (e), solid lines show the fitted relationship between objective value and solving time, and the accompanying labels report the corresponding slopes. Gray dashed lines mark the mean objective value and solving time at each stage, with labels reporting the proportion of instances below the mean objective value. Orange points identify a low-objective, high-time subset at iteration 10 and track the same instances across all three checkpoints.}
\label{fig:observation-trace}
\label{fig:mechanistic-observations}
\label{fig:source-imbalance}
\end{figure*}

\section{Preliminary}
\label{sec:preliminary}

\begingroup

\paragraph{LLM-Based Automatic Heuristic Design.}
Given a prompt specifying the task and required function interface, the LLM iteratively generates or revises candidate heuristics using selected parents and their evaluation feedback~\citep{liu2024eoh,ye2024reevo}. Each candidate heuristic is embedded in a fixed search framework to form a heuristic solver for tour optimization~\citep{voudouris1999gls}. The resulting solver is evaluated on a set of instances, and its performance guides candidate selection and subsequent generation to evolve effective heuristics.

\looseness=-1 Formally, let $D_s=\{x_i\}_{i=1}^{N}$ denote the source instance set, where $N=1$ corresponds to single-instance evaluation. Given a candidate heuristic $h$, the search framework uses $h$ to obtain a tour for each instance in $D_s$, and the task-specific evaluation $\mathbf{E}(h;D_s)$ summarizes its performance. At iteration $t$, let $\mathcal{P}_t$ denote the current population and $\mathcal{Q}_t$ the set of newly generated or revised candidate heuristics. The population is updated as
\begin{equation}
\label{eq:population-update}
\mathcal{P}_{t+1}
=
\operatorname{Select}_{M}
\left(
\mathcal{P}_t\cup\mathcal{Q}_t;
\mathbf{E}(\cdot;D_s)
\right),
\end{equation}
where $\operatorname{Select}_{M}$ retains $M$ candidate heuristics according to their evaluation results. The selected population and its evaluation results jointly guide candidate generation in the next iteration.

\paragraph{Dynamic Re-solving.}
 When location changes, new requests, or request cancellations alter the set of unserved requests or their locations during tour execution, the remaining task must be reoptimized. The executed prefix remains fixed, while the remaining task requires a path from the current location through the updated unvisited nodes to the original starting node. The search framework starts from a warm-start route, initializes its search state, and reuses the heuristic previously evolved on static instances to optimize the remaining task.

Formally, let $h^\star$ denote the heuristic selected after evolution on the source instance set $D_s$, $\pi_i$ the execution tour for source instance $x_i$, and $\Delta_i$ the change event at tour progress $\rho$. The corresponding residual instance is
\begin{equation}
 x_{i,\rho}^{\mathrm{res}}=R(x_i,\pi_i,\rho,\Delta_i),
\end{equation}
where $R$ constructs the remaining task from the execution state and updated node information. The set $D_\rho=\{x_{i,\rho}^{\mathrm{res}}\}_{i=1}^{N}$ forms a target instance set induced by route execution. The evolved heuristic $h^\star$ is used to solve these residual instances to evaluate its performance after dynamic changes.
\endgroup

\section{Observation}
\label{sec:observation}

The following analysis first investigates why re-solving updated instances becomes inefficient and then examines the key factors in solver evolution that give rise to this behavior.

\subsection{Observation 1. Why Inefficient?}
\label{sec:observation-overfitted-search}
\looseness=-1 To examine why solving time increases when re-solving updated instances in Figure~\ref{fig:observation}, we analyze the solver's search behavior during this process. We take the best EoH heuristic evolved on static TSP200 instances and reuse it within GLS to re-solve the remaining task after 20\% tour progress, with new nodes added at 20\% of the unvisited-node count. Figure~\ref{fig:observation-trace}~(a) shows the edge selections on the original 200-node instance, while Figure~\ref{fig:observation-trace}~(b) shows the edge selections on the changed instance, with thicker lines indicating higher selection frequencies. These edges concentrate in a small region, indicating that the search repeatedly revisits overlapping local neighborhoods. This behavior suggests that ``static-overfitting'' may contribute to re-solving inefficiency, as static evolution biases the heuristic toward structural patterns in the source instances and may thereby increase the search effort required to obtain high-quality solutions after instance changes. The evolved solver may thus become specialized to the static scenario and struggle to maintain stable search behavior after instance changes.

\begin{figure*}[!tbp]
\centering
\includegraphics[width=0.98\textwidth,pagebox=cropbox]{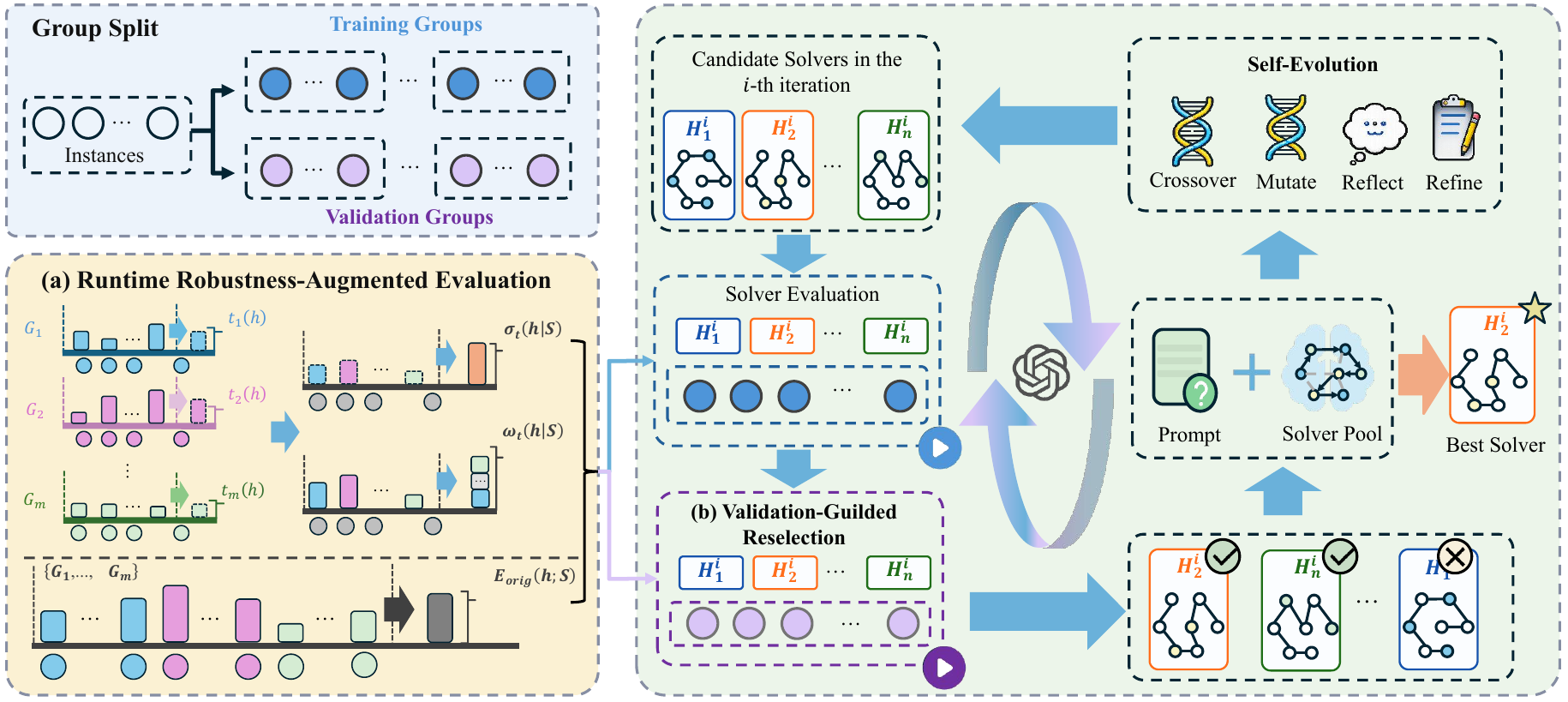}
\caption{Overview of the \method{} framework. Training and validation groups support solver evaluation and validation-guided reselection, respectively. (a) The original objective used in solver evaluation and reselection is augmented with inter-group variability and tail risk in solving time. (b) Validation-guided reselection reevaluates candidates to update the solver pool before the next evolution.}
\label{fig:tare-framework-overview}
\end{figure*}

\subsection{Observation 2. How ``Overfitting'' Emerges?}
\label{sec:observation-source-imbalance}
\looseness=-1 We investigate how ``static-overfitting'' emerges by tracking the solvers retained by EoH at different stages of evolution on TSP200. At different iterations, we evaluate the evolved solver obtained at each checkpoint on the same 50 training instances. For each instance, we record the solving time and objective value. Figure~\ref{fig:mechanistic-observations}~(c)--(e) visualizes the joint distributions of these two measures across the three checkpoints and reveals two observations: 1) Across the three checkpoints, clear quality improvements appear in an increasingly small subset of training instances, while most instances show only minor changes or even quality degradation; 2) larger quality improvements are observed mainly on instances requiring longer solving times, and this pattern becomes progressively clearer across iterations. These results show that existing evolutionary pattern weakly penalizes solvers whose gains are concentrated on a narrow subset of instances and accompanied by increasing search cost, allowing them to remain competitive during evolution.

\section{Methodology}
\label{sec:methodology}

\looseness=-1 Motivated by the two deficiencies identified above, \method{} augments existing AHD frameworks with validation-guided reselection and runtime robustness-augmented evaluation to reduce solver ``overfitting'' to fixed training instances. Validation-guided reselection reevaluates candidate solvers on a separate set excluded from routine evolution, reducing specialization to the instances repeatedly used during evolution. Inter-group variability and tail risk are further incorporated into the evaluation objectives, encouraging candidate selection to account for search stability.

\subsection{Validation-Guided Reselection}

Drawing on the train--validation split used to curb overfitting in model selection \citep{cawley2010overfitting}, we divide the instances into disjoint training and validation groups for routine evolution and reselection, respectively. Given a source instance set $D_s$, we randomly partition it into $K$ groups, denoted by $\mathcal{B}=\{\mathcal{G}_1,\ldots,\mathcal{G}_K\}$, and assign them to the training split $\mathcal{B}_{\train}$ and validation split $\mathcal{B}_{\val}$ such that
\begin{equation}
\mathcal{B}=\mathcal{B}_{\train}\cup\mathcal{B}_{\val}.
\end{equation}
\begin{equation}
\mathcal{B}_{\train}\cap\mathcal{B}_{\val}=\emptyset.
\end{equation}

New candidate solvers generated from the current population are initially evaluated on $\mathcal{B}_{\train}$.  Whenever the underlying AHD method updates its population, we evaluate the candidates entering survivor selection on $\mathcal{B}_{\val}$ under the same criteria and use their validation results to select survivors. 
After evolution, we select the heuristic with the highest DDS score on $\mathcal{B}_{\val}$ from the final population for dynamic testing.

\subsection{Runtime Robustness-Augmented Evaluation}

\looseness=-1 Solver performance can vary substantially across problem instances, a central premise of algorithm selection research~\citep{rice1976algorithm}. We therefore introduce two criteria, inter-group variability and tail risk in solving-time, to penalize candidates that induce uneven or excessive search effort across instance groups. These criteria augment the original evaluation objective of each integrated AHD method, allowing runtime robustness to inform candidate selection. Formally, let $\mathcal{S}=\{\mathcal{G}_1,\ldots,\mathcal{G}_m\}$ denote the groups under evaluation and $t_q(h)$ denote the mean time required to solve a tour with the solver using heuristic $h$ on the instances in $\mathcal{G}_q$. We define inter-group variability as
\begin{equation}
\sigma_t(h\mid\mathcal{S})=
\left(
\frac{1}{m}\sum_{q=1}^{m}
\left(
t_q(h)-\frac{1}{m}\sum_{r=1}^{m}t_r(h)
\right)^2
\right)^{1/2},
\end{equation}
which penalizes candidates whose tour solving time varies sharply across groups, thereby favoring stable search efficiency across different instance subsets.

Motivated by the upper-tail aggregation principle underlying conditional value-at-risk \citep{rockafellar2000cvar}, we define tail risk by ordering the group times as $t_{(1)}(h)\ge\cdots\ge t_{(m)}(h)$ and setting $r=\lceil m/2\rceil$:
\begin{equation}
w_t(h\mid\mathcal{S})=
\frac{1}{r}\sum_{q=1}^{r}t_{(q)}(h),
\end{equation}
which penalizes high solving times among the slower groups.

Let $\mathbf{E}_{\mathrm{orig}}(h;\mathcal{S})$ denote the original evaluation objectives of the integrated AHD method, including its conventional solution-quality objective and any method-specific objectives. \method{} supplements these objectives with inter-group variability and tail risk as follows.
\begin{equation}
\mathbf{z}(h\mid\mathcal{S})
=
\left[
\mathbf{E}_{\mathrm{orig}}(h;\mathcal{S})^\top,
\sigma_t(h\mid\mathcal{S}),
w_t(h\mid\mathcal{S})
\right]^\top.
\end{equation}

The augmented objectives preserve the original criteria while explicitly accounting for inter-group variability and solving times in the slower groups during candidate comparison. The two additional criteria incorporate the associated computational costs into evolutionary selection.

Assuming that all objectives are expressed in minimization form, candidates are ranked by Pareto dominance over the augmented objectives \citep{deb2002nsga2,emmerich2018multiobjective}. For two candidates $h_i$ and $h_j$, we define
\begin{equation}
\begin{aligned}
h_i \prec_{\mathcal{S}} h_j
\Longleftrightarrow\;&
\mathbf{z}(h_i\mid\mathcal{S})\le
\mathbf{z}(h_j\mid\mathcal{S})\\
&\land\
\mathbf{z}(h_i\mid\mathcal{S})\ne
\mathbf{z}(h_j\mid\mathcal{S}),
\end{aligned}
\end{equation}
where the inequality is interpreted componentwise. The relation $h_i \prec_{\mathcal{S}} h_j$ indicates that $h_i$ Pareto-dominates $h_j$: $h_i$ is no worse on any criterion and strictly better on at least one. 

Survivor selection prioritizes nondominated candidates and ranks candidates within each category by descending dominance--dissimilarity scores (DDS) following MEoH~\citep{yao2025meoh}. DDS ties are resolved by the integrated AHD method's original ranking rule, followed by tail risk and inter-group variability. Candidates are retained in this order up to the population budget.

\subsection{Framework Overview}

\method{} is designed as a plug-in framework for solver evolution, and its overall workflow is illustrated in Figure~\ref{fig:tare-framework-overview}. Starting from an initial population, the underlying AHD method generates new candidate solvers through its original evolutionary operators. \method{} evaluates the expanded candidate pool under the proposed selection strategy and retains promising solvers for the next generation.  At each population update, candidates entering survivor selection are evaluated on the validation instances, and their validation results determine the survivors before evolution continues.

% Experimental floats use standard IEEEtran placement.
\section{Experiments}
\label{sec:experiments}

\subsection{Datasets}
\looseness=-1 We generate 50 synthetic Euclidean TSP instances per size $n\in\{200,500\}$, with node coordinates sampled independently and uniformly from $[0,1]^2$, following EoH~\citep{liu2024eoh} and ReEvo~\citep{ye2024reevo}. For each size, we randomly partition the instances into 10 equal groups, using a default 5:5 split between routine evolution and validation-guided reselection (fixed per run). We also conduct dynamic tests on TSPLIB~\citep{reinelt1991tsplib}, a standard TSP benchmark library containing real-world application instances (Section~\ref{sec:tsplib-generalization}).

\subsection{Dynamic Scenario}
\label{app:experimental-details}

We evaluate nine dynamic TSP scenarios on all 50 source instances, combining the three change types in Table~\ref{tab:dynamic-construction} with 20\%, 50\%, and 80\% tour progress. Location changes modify the coordinate data of existing requests, while new requests and request cancellations add or remove requests from the unserved set. We use $\sigma=0.02$ and $r_1=r_2=20\%$ by default.
For an original tour $\pi=(v_1,\ldots,v_n)$, we trigger a dynamic change after visiting a fraction $\rho\in(0,1)$ of the nodes. Counting the starting node, the executed prefix contains
$k_{\rho}=\lceil \rho n\rceil$
visited nodes, with $1\leq k_{\rho}\leq n-1$. We apply changes in Table~\ref{tab:dynamic-construction} only to the unvisited set $U_{\rho}=\{v_{k_{\rho}+1},\ldots,v_n\}$, keeping the executed prefix fixed. We reoptimize the path from the current location $s=v_{k_{\rho}}$ through the changed set $U'_{\rho}$ to the original start $t=v_1$, adding a dummy node $d$ to form a closed-tour TSP with edge costs
\begin{equation}
 c(d,u)=c(u,d)=
 \begin{cases}
  0, & u\in\{s,t\},\\
  M, & u\in U'_{\rho},
 \end{cases}
\end{equation}
\looseness=-1 where $M$ is a large finite penalty exceeding the cost of a feasible $s$-to-$t$ path through all remaining nodes. We preserve real-node distances and require $s$ and $t$ to be the two neighbors of $d$ in the returned tour. We exclude dummy edges from distance statistics and heuristic updates, retaining their prescribed costs throughout search. The zero-cost links $t\!\to\!d\!\to\!s$ close the path, and removing $d$ recovers an equal-cost $s$-to-$t$ route.

\begin{table}[!htbp]
\renewcommand{\arraystretch}{0.85}
\centering
\caption{Construction of the three dynamic changes. Each operation is applied to the unvisited set after the tour-progress cutoff.}
\label{tab:dynamic-construction}

\begingroup
\footnotesize
\setlength{\tabcolsep}{3pt}
\begin{tabular}{>{\raggedright\arraybackslash}p{0.18\columnwidth}p{0.77\columnwidth}}
\toprule
Change & Operation \\
\midrule
Location changes & Add independent zero-mean Gaussian noise with standard deviation $\sigma$ to both coordinates of every node in $U_{\rho}$, then clip the perturbed coordinates to $[0,1]^2$. The node count remains unchanged. \\
\midrule
New requests & Add $\max\{1,\operatorname{round}(r_1|U_{\rho}|)\}$ nodes sampled independently and uniformly from $[0,1]^2$ to the unvisited set, where $r_1$ denotes the node-addition ratio. \\
\midrule
Request cancellations & Remove $\max\{1,\operatorname{round}(r_2|U_{\rho}|)\}$ nodes sampled uniformly without replacement from the unvisited set, where $r_2$ denotes the node-removal ratio. The removal count is capped at $|U_{\rho}|-2$, and retained-node coordinates remain unchanged. \\
\bottomrule
\end{tabular}
\endgroup

\end{table}

For each source instance, we sample events during execution of a Concorde-generated static tour~\citep{applegate2006tsp} to construct shared dynamic instances. We warm-start all methods from the same adapted remaining tour, preserving visiting order for location changes, removing canceled nodes, and sequentially inserting nodes for new requests where they least increase route length.

\flushbottom
\subsection{Evaluation Metrics}

We measure objective value (Obj.) and solving time. Obj. is the total length of the returned route, computed using the original distance matrix. In dynamic evaluation, it measures the remaining route from the current location through the updated unvisited nodes to the original start. We report mean objective values over the evaluated instances within each scenario.
In both evolution and dynamic evaluation, we measure wall-clock search time from initiation to termination under the specified stopping criteria, including heuristic calls and search operations but excluding initial-route construction. Lower objective values and shorter times are better.

\subsection{Baselines}
We pair four representative AHD methods, EoH~\citep{liu2024eoh}, ReEvo~\citep{ye2024reevo}, Hercules~\citep{wu2025hercules}, and MEoH~\citep{yao2025meoh}, with their \method{} variants, preserving solver generation and evolutionary schedules while adding runtime-robust evaluation and validation-guided reselection to candidate selection.

\begin{table*}[!tbp]
\renewcommand{\arraystretch}{0.85}
\centering
\caption{Mean solving time (s) on TSP200 and TSP500 under dynamic scenarios within GLS and ILS. The column groups 20\%, 50\%, and 80\% indicate the fraction of nodes visited when a dynamic change occurs. Loc. Change, New Req., and Cancel. denote location changes, new requests, and request cancellations, respectively. For each scenario, solving times are first averaged over paired successful instances in each run. We report the mean $\pm$ standard deviation of these instance-averaged times across three runs. Within each baseline pair, the result with the lower mean for each scale and scenario is shown in bold, including its standard deviation.}
\label{tab:repo-large-scale}
\label{tab:repo-tsp200}
\label{tab:repo-tsp500}
\label{tab:ils-time}
\begingroup
\footnotesize
\setlength{\tabcolsep}{0pt}
\begin{tabular*}{\textwidth}{@{\extracolsep{\fill}}ll*{9}{c}@{}}
\toprule
 & \textbf{Method} & \multicolumn{3}{c}{\textbf{20\%}} & \multicolumn{3}{c}{\textbf{50\%}} & \multicolumn{3}{c}{\textbf{80\%}}\\
\cmidrule(lr){3-5}\cmidrule(lr){6-8}\cmidrule(l){9-11}
& & Loc. Change\,$\downarrow$ & New Req.\,$\downarrow$ & Cancel.\,$\downarrow$ & Loc. Change\,$\downarrow$ & New Req.\,$\downarrow$ & Cancel.\,$\downarrow$ & Loc. Change\,$\downarrow$ & New Req.\,$\downarrow$ & Cancel.\,$\downarrow$\\
\specialrule{0.4pt}{2pt}{0pt}
\rowcolor{gray!15}\multicolumn{11}{@{}l@{}}{\strut\textbf{GLS}}\\
\specialrule{0.4pt}{0pt}{2pt}
& EoH & \mstd{6.443}{0.709} & \mstd{9.139}{1.106} & \mstd{4.412}{0.437} & \mstd{2.757}{0.276} & \mstd{4.025}{0.443} & \mstd{1.860}{0.167} & \mstd{0.588}{0.053} & \mstd{0.782}{0.077} & \mstd{0.443}{0.036}\\
 & \hspace{0.5em}\textbf{+\method{}} & \textbf{\boldmath\mstd{2.673}{0.176}} & \textbf{\boldmath\mstd{3.806}{0.276}} & \textbf{\boldmath\mstd{2.547}{0.151}} & \textbf{\boldmath\mstd{0.443}{0.027}} & \textbf{\boldmath\mstd{0.508}{0.034}} & \textbf{\boldmath\mstd{0.361}{0.019}} & \textbf{\boldmath\mstd{0.220}{0.012}} & \textbf{\boldmath\mstd{0.242}{0.014}} & \textbf{\boldmath\mstd{0.189}{0.009}}\\
\addlinespace[0.8pt]
 & MEoH & \mstd{1.420}{0.209} & \mstd{2.009}{0.070} & \mstd{0.930}{0.054} & \mstd{0.582}{0.041} & \mstd{0.822}{0.073} & \mstd{0.381}{0.014} & \mstd{0.102}{0.006} & \mstd{0.144}{0.010} & \mstd{0.067}{0.004}\\
 & \hspace{0.5em}\textbf{+\method{}} & \textbf{\boldmath\mstd{0.986}{0.039}} & \textbf{\boldmath\mstd{1.349}{0.073}} & \textbf{\boldmath\mstd{0.672}{0.042}} & \textbf{\boldmath\mstd{0.439}{0.045}} & \textbf{\boldmath\mstd{0.601}{0.061}} & \textbf{\boldmath\mstd{0.299}{0.012}} & \textbf{\boldmath\mstd{0.091}{0.004}} & \textbf{\boldmath\mstd{0.124}{0.006}} & \textbf{\boldmath\mstd{0.062}{0.002}}\\
\addlinespace[0.8pt]
 & ReEvo & \mstd{1.346}{0.123} & \mstd{1.954}{0.230} & \mstd{0.880}{0.096} & \mstd{0.535}{0.059} & \mstd{0.783}{0.075} & \mstd{0.367}{0.036} & \mstd{0.109}{0.021} & \mstd{0.153}{0.012} & \mstd{0.078}{0.007}\\
 & \hspace{0.5em}\textbf{+\method{}} & \textbf{\boldmath\mstd{0.566}{0.054}} & \textbf{\boldmath\mstd{0.822}{0.052}} & \textbf{\boldmath\mstd{0.373}{0.016}} & \textbf{\boldmath\mstd{0.236}{0.007}} & \textbf{\boldmath\mstd{0.336}{0.026}} & \textbf{\boldmath\mstd{0.161}{0.010}} & \textbf{\boldmath\mstd{0.049}{0.003}} & \textbf{\boldmath\mstd{0.068}{0.005}} & \textbf{\boldmath\mstd{0.035}{0.002}}\\
\addlinespace[0.8pt]
 & Hercules & \mstd{1.079}{0.059} & \mstd{1.553}{0.094} & \mstd{0.691}{0.034} & \mstd{0.428}{0.021} & \mstd{0.614}{0.034} & \mstd{0.282}{0.013} & \mstd{0.079}{0.004} & \mstd{0.111}{0.005} & \mstd{0.054}{0.002}\\
\multirow{-8}{*}{\rotatebox[origin=c]{90}{TSP200}} & \hspace{0.5em}\textbf{+\method{}} & \textbf{\boldmath\mstd{0.432}{0.017}} & \textbf{\boldmath\mstd{0.622}{0.026}} & \textbf{\boldmath\mstd{0.279}{0.010}} & \textbf{\boldmath\mstd{0.180}{0.006}} & \textbf{\boldmath\mstd{0.251}{0.010}} & \textbf{\boldmath\mstd{0.118}{0.004}} & \textbf{\boldmath\mstd{0.034}{0.002}} & \textbf{\boldmath\mstd{0.047}{0.002}} & \textbf{\boldmath\mstd{0.023}{0.002}}\\
\midrule
 & EoH & \mstd{10.014}{0.218} & \mstd{10.018}{0.304} & \mstd{5.552}{0.262} & \mstd{3.224}{0.646} & \mstd{4.987}{1.033} & \mstd{2.171}{0.369} & \mstd{0.516}{0.101} & \mstd{0.896}{0.147} & \mstd{0.401}{0.064}\\
 & \hspace{0.5em}\textbf{+\method{}} & \textbf{\boldmath\mstd{2.447}{0.334}} & \textbf{\boldmath\mstd{3.134}{0.532}} & \textbf{\boldmath\mstd{1.476}{0.191}} & \textbf{\boldmath\mstd{1.031}{0.104}} & \textbf{\boldmath\mstd{1.578}{0.155}} & \textbf{\boldmath\mstd{0.793}{0.066}} & \textbf{\boldmath\mstd{0.414}{0.028}} & \textbf{\boldmath\mstd{0.489}{0.037}} & \textbf{\boldmath\mstd{0.348}{0.019}}\\
\addlinespace[0.8pt]
 & MEoH & \mstd{8.101}{0.824} & \mstd{10.436}{0.171} & \mstd{5.301}{0.667} & \mstd{3.317}{0.432} & \mstd{4.690}{0.361} & \mstd{2.171}{0.137} & \mstd{0.582}{0.037} & \mstd{0.822}{0.157} & \mstd{0.381}{0.029}\\
 & \hspace{0.5em}\textbf{+\method{}} & \textbf{\boldmath\mstd{5.270}{0.261}} & \textbf{\boldmath\mstd{7.212}{0.393}} & \textbf{\boldmath\mstd{3.591}{0.160}} & \textbf{\boldmath\mstd{2.348}{0.106}} & \textbf{\boldmath\mstd{3.213}{0.159}} & \textbf{\boldmath\mstd{1.600}{0.065}} & \textbf{\boldmath\mstd{0.486}{0.020}} & \textbf{\boldmath\mstd{0.664}{0.030}} & \textbf{\boldmath\mstd{0.331}{0.012}}\\
\addlinespace[0.8pt]
 & ReEvo & \mstd{1.853}{0.465} & \mstd{1.700}{0.651} & \mstd{0.626}{0.262} & \mstd{0.453}{0.147} & \mstd{0.573}{0.230} & \mstd{0.218}{0.072} & \mstd{0.087}{0.024} & \mstd{0.096}{0.044} & \mstd{0.046}{0.020}\\
 & \hspace{0.5em}\textbf{+\method{}} & \textbf{\boldmath\mstd{0.668}{0.146}} & \textbf{\boldmath\mstd{0.921}{0.236}} & \textbf{\boldmath\mstd{0.399}{0.082}} & \textbf{\boldmath\mstd{0.263}{0.050}} & \textbf{\boldmath\mstd{0.357}{0.080}} & \textbf{\boldmath\mstd{0.168}{0.030}} & \textbf{\boldmath\mstd{0.053}{0.010}} & \textbf{\boldmath\mstd{0.076}{0.014}} & \textbf{\boldmath\mstd{0.035}{0.006}}\\
\addlinespace[0.8pt]
 & Hercules & \mstd{4.341}{0.239} & \mstd{6.355}{0.484} & \mstd{2.847}{0.141} & \mstd{1.717}{0.086} & \mstd{2.546}{0.110} & \mstd{1.104}{0.050} & \mstd{0.304}{0.018} & \mstd{0.422}{0.041} & \mstd{0.188}{0.008}\\
\multirow{-8}{*}{\rotatebox[origin=c]{90}{TSP500}} & \hspace{0.5em}\textbf{+\method{}} & \textbf{\boldmath\mstd{2.045}{0.079}} & \textbf{\boldmath\mstd{3.489}{0.148}} & \textbf{\boldmath\mstd{1.345}{0.047}} & \textbf{\boldmath\mstd{0.878}{0.031}} & \textbf{\boldmath\mstd{1.342}{0.052}} & \textbf{\boldmath\mstd{0.697}{0.022}} & \textbf{\boldmath\mstd{0.188}{0.006}} & \textbf{\boldmath\mstd{0.200}{0.007}} & \textbf{\boldmath\mstd{0.090}{0.003}}\\
\specialrule{0.4pt}{2pt}{0pt}
\rowcolor{gray!15}\multicolumn{11}{@{}l@{}}{\strut\textbf{ILS}}\\
\specialrule{0.4pt}{0pt}{2pt}
& EoH & \mstd{3.421}{0.431} & \mstd{4.441}{0.252} & \mstd{2.609}{0.206} & \mstd{2.044}{0.118} & \mstd{2.455}{0.238} & \mstd{1.738}{0.102} & \mstd{1.140}{0.192} & \mstd{1.197}{0.065} & \mstd{1.030}{0.031}\\
 & \hspace{0.5em}\textbf{+\method{}} & \textbf{\boldmath\mstd{0.370}{0.044}} & \textbf{\boldmath\mstd{0.525}{0.061}} & \textbf{\boldmath\mstd{0.313}{0.007}} & \textbf{\boldmath\mstd{0.220}{0.063}} & \textbf{\boldmath\mstd{0.284}{0.066}} & \textbf{\boldmath\mstd{0.175}{0.041}} & \textbf{\boldmath\mstd{0.080}{0.002}} & \textbf{\boldmath\mstd{0.095}{0.006}} & \textbf{\boldmath\mstd{0.064}{0.001}}\\
\addlinespace[0.8pt]
 & MEoH & \mstd{0.277}{0.041} & \mstd{0.354}{0.046} & \mstd{0.208}{0.008} & \mstd{0.148}{0.007} & \mstd{0.170}{0.012} & \mstd{0.099}{0.024} & \mstd{0.050}{0.003} & \mstd{0.073}{0.004} & \mstd{0.046}{0.008}\\
 & \hspace{0.5em}\textbf{+\method{}}  & \textbf{\boldmath\mstd{0.155}{0.020}} & \textbf{\boldmath\mstd{0.189}{0.007}} & \textbf{\boldmath\mstd{0.120}{0.021}} & \textbf{\boldmath\mstd{0.090}{0.010}} & \textbf{\boldmath\mstd{0.100}{0.020}} & \textbf{\boldmath\mstd{0.062}{0.005}} & \textbf{\boldmath\mstd{0.036}{0.002}} & \textbf{\boldmath\mstd{0.051}{0.006}} & \textbf{\boldmath\mstd{0.034}{0.004}}\\
\addlinespace[0.8pt]
 & ReEvo & \mstd{6.787}{1.140} & \mstd{8.956}{0.884} & \mstd{6.112}{0.593} & \mstd{4.642}{0.903} & \mstd{5.701}{0.252} & \mstd{3.935}{0.234} & \mstd{2.012}{0.349} & \mstd{2.447}{0.218} & \mstd{1.670}{0.037}\\
 & \hspace{0.5em}\textbf{+\method{}} & \textbf{\boldmath\mstd{1.005}{0.031}} & \textbf{\boldmath\mstd{1.406}{0.066}} & \textbf{\boldmath\mstd{0.735}{0.128}} & \textbf{\boldmath\mstd{0.534}{0.035}} & \textbf{\boldmath\mstd{0.689}{0.026}} & \textbf{\boldmath\mstd{0.395}{0.052}} & \textbf{\boldmath\mstd{0.187}{0.036}} & \textbf{\boldmath\mstd{0.224}{0.008}} & \textbf{\boldmath\mstd{0.147}{0.016}}\\
\addlinespace[0.8pt]
 & Hercules & \mstd{0.973}{0.066} & \mstd{1.449}{0.118} & \mstd{0.679}{0.125} & \mstd{0.436}{0.065} & \mstd{0.607}{0.012} & \mstd{0.304}{0.024} & \mstd{0.109}{0.019} & \mstd{0.141}{0.020} & \mstd{0.085}{0.011}\\
\multirow{-8}{*}{\rotatebox[origin=c]{90}{TSP200}} & \hspace{0.5em}\textbf{+\method{}} & \textbf{\boldmath\mstd{0.288}{0.022}} & \textbf{\boldmath\mstd{0.388}{0.020}} & \textbf{\boldmath\mstd{0.197}{0.012}} & \textbf{\boldmath\mstd{0.137}{0.011}} & \textbf{\boldmath\mstd{0.186}{0.021}} & \textbf{\boldmath\mstd{0.099}{0.012}} & \textbf{\boldmath\mstd{0.043}{0.004}} & \textbf{\boldmath\mstd{0.052}{0.006}} & \textbf{\boldmath\mstd{0.033}{0.006}}\\
\midrule
 & EoH  & \mstd{8.536}{0.362} & \mstd{8.991}{1.234} & \mstd{6.594}{0.425} & \mstd{4.825}{0.376} & \mstd{6.261}{0.442} & \mstd{4.128}{0.613} & \mstd{2.165}{0.330} & \mstd{2.728}{0.097} & \mstd{1.913}{0.053}\\
 & \hspace{0.5em}\textbf{+\method{}}  & \textbf{\boldmath\mstd{1.388}{0.181}} & \textbf{\boldmath\mstd{1.729}{0.169}} & \textbf{\boldmath\mstd{1.178}{0.049}} & \textbf{\boldmath\mstd{0.895}{0.102}} & \textbf{\boldmath\mstd{1.198}{0.083}} & \textbf{\boldmath\mstd{0.792}{0.086}} & \textbf{\boldmath\mstd{0.514}{0.028}} & \textbf{\boldmath\mstd{0.568}{0.069}} & \textbf{\boldmath\mstd{0.444}{0.047}}\\
\addlinespace[0.8pt]
 & MEoH  & \mstd{1.555}{0.175} & \mstd{1.997}{0.091} & \mstd{1.127}{0.133} & \mstd{0.771}{0.122} & \mstd{0.932}{0.121} & \mstd{0.519}{0.047} & \mstd{0.242}{0.030} & \mstd{0.371}{0.042} & \mstd{0.214}{0.002}\\
 & \hspace{0.5em}\textbf{+\method{}}  & \textbf{\boldmath\mstd{1.015}{0.041}} & \textbf{\boldmath\mstd{1.216}{0.062}} & \textbf{\boldmath\mstd{0.791}{0.108}} & \textbf{\boldmath\mstd{0.556}{0.083}} & \textbf{\boldmath\mstd{0.610}{0.058}} & \textbf{\boldmath\mstd{0.391}{0.042}} & \textbf{\boldmath\mstd{0.198}{0.023}} & \textbf{\boldmath\mstd{0.286}{0.012}} & \textbf{\boldmath\mstd{0.182}{0.007}}\\
\addlinespace[0.8pt]
 & ReEvo & \mstd{2.031}{0.045} & \mstd{2.875}{0.057} & \mstd{1.250}{0.128} & \mstd{0.776}{0.104} & \mstd{1.158}{0.183} & \mstd{0.599}{0.065} & \mstd{0.210}{0.024} & \mstd{0.305}{0.028} & \mstd{0.162}{0.002}\\
 & \hspace{0.5em}\textbf{+\method{}} & \textbf{\boldmath\mstd{1.728}{0.025}} & \textbf{\boldmath\mstd{2.337}{0.224}} & \textbf{\boldmath\mstd{0.946}{0.106}} & \textbf{\boldmath\mstd{0.607}{0.053}} & \textbf{\boldmath\mstd{0.838}{0.102}} & \textbf{\boldmath\mstd{0.399}{0.043}} & \textbf{\boldmath\mstd{0.138}{0.012}} & \textbf{\boldmath\mstd{0.175}{0.003}} & \textbf{\boldmath\mstd{0.098}{0.011}}\\
\addlinespace[0.8pt]
 & Hercules & \mstd{5.372}{0.203} & \mstd{8.881}{1.108} & \mstd{3.498}{0.161} & \mstd{2.078}{0.442} & \mstd{3.196}{0.245} & \mstd{1.469}{0.091} & \mstd{0.448}{0.029} & \mstd{0.639}{0.071} & \mstd{0.324}{0.025}\\
\multirow{-8}{*}{\rotatebox[origin=c]{90}{TSP500}} & \hspace{0.5em}\textbf{+\method{}} & \textbf{\boldmath\mstd{1.927}{0.229}} & \textbf{\boldmath\mstd{2.564}{0.134}} & \textbf{\boldmath\mstd{1.086}{0.171}} & \textbf{\boldmath\mstd{0.719}{0.096}} & \textbf{\boldmath\mstd{0.989}{0.051}} & \textbf{\boldmath\mstd{0.490}{0.053}} & \textbf{\boldmath\mstd{0.186}{0.033}} & \textbf{\boldmath\mstd{0.230}{0.003}} & \textbf{\boldmath\mstd{0.144}{0.021}}\\
\bottomrule
\end{tabular*}
\endgroup
\end{table*}

\subsection{Search Frameworks and Operators}
\label{sec:search-frameworks}

\looseness=-1 We evaluate \method{} within two search frameworks, GLS and iterated local search (ILS), in our experiments. Within GLS~\citep{voudouris1999gls}, we evolve \texttt{update\_\allowbreak edge\_\allowbreak distance} to map the original distance matrix, current tour, and edge-use counts accumulated during perturbation to modified distances guiding 2-opt and relocate perturbations. A 2-opt move reverses the segment between two removed tour edges and reconnects the tour, while relocate reinserts a node elsewhere.
\looseness=-1 Within ILS~\citep{lourenco2010ils}, we evolve \texttt{perturb\_\allowbreak tour} to map the original distance matrix and current tour to a perturbed tour with the same node set. We use double-bridge perturbations to cut four tour edges and reconnect the resulting segments in a different order. In both frameworks, we improve perturbed tours with fixed 2-opt and relocate local search under the original distances. For ILS, we accept a shorter tour and start the next search iteration from the newly accepted \mbox{tour as usual}.

\subsection{Experimental Setup}
\label{sec:experimental-setup}

\looseness=-1 Experiments run locally on a 64-bit Windows 11 Pro machine with an Intel Core i5-12500 CPU (6 cores, 12 threads) and 16\,GB RAM. We evolve heuristics separately within each search framework using gpt-6-luna. GLS and ILS share a limit of 500 iterations and a per-instance time budget of 10 seconds for both TSP200 and TSP500, stopping when either limit is reached. We check time before each iteration, allow an ongoing iteration to finish, and return the best tour found. GLS uses two perturbation moves per iteration. We use identical search settings and evaluation seeds within each baseline pair. We run EoH for ten generations with five operators producing four offspring each. For Hercules and ReEvo, we use candidate-generation budgets of 50 and 100, respectively, with populations of four and initial populations of ten. We cap MEoH at 220 sampled candidates, with four samplers and four evaluators.  By default, validation-guided reselection is performed whenever the underlying AHD method updates its population.

\paragraph{Failure Handling during Evolution.}
We assign failed candidate evaluations an infeasible quality penalty and a time penalty of $2T$, where $T$ is the per-instance search budget. We evaluate candidates in a separate process with a batch timeout of $\max(300,NT+15)$ seconds for $N$ instances, excluding recorded startup, data-loading, and initialization time. We terminate timed-out batches and reject their candidates.

\paragraph{Failure Handling during Dynamic Evaluation.}
We count execution errors, invalid tours, non-finite costs, and watchdog termination as failures, and budget exhaustion with a valid tour as success. For each baseline--ERDR comparison, we average search times and objective values over the same instances on which both methods succeed. Across the evaluated comparisons, at least 46 of the 50 instance pairs were retained, with the largest observed exclusion count being four.

\lstdefinestyle{staticcode}{
  language=Python,
  basicstyle=\ttfamily\footnotesize,
  keywordstyle=\color{black},
  commentstyle=\color{orange!85!black}\bfseries,
  columns=fullflexible,
  keepspaces=true,
  showstringspaces=false,
  breaklines=true,
  breakatwhitespace=false,
  backgroundcolor=\color{orange!6},
  frame=single,
  rulecolor=\color{orange!45},
  framesep=3pt,
  aboveskip=3pt,
  belowskip=2pt
}
\lstdefinestyle{erdrcode}{
  language=Python,
  basicstyle=\ttfamily\footnotesize,
  keywordstyle=\color{black},
  commentstyle=\color{green!45!black}\bfseries,
  columns=fullflexible,
  keepspaces=true,
  showstringspaces=false,
  breaklines=true,
  breakatwhitespace=false,
  backgroundcolor=\color{green!6},
  frame=single,
  rulecolor=\color{green!45!black},
  framesep=3pt,
  aboveskip=3pt,
  belowskip=2pt
}

\begin{table*}[!tbp]
\renewcommand{\arraystretch}{0.85}
\centering
\caption{Mean final objective value (Obj.) on TSP200 and TSP500 under dynamic scenarios within GLS and ILS. For each scenario, objective values are first averaged over paired successful instances in each run, and the reported value is the mean of these instance averages across three runs. Scenario columns follow Table~\ref{tab:repo-large-scale}. $\Delta$ annotations indicate the objective-value differences of ERDR relative to the corresponding baselines. Lower values are better.}
\label{tab:objective-results}
\label{tab:gap-difference}
\label{tab:ils-gap}
\begingroup
\footnotesize
\setlength{\tabcolsep}{0pt}
\newcommand{\objmean}[1]{\makebox[45pt][c]{{\scriptsize #1}}}
\def\objdeltasigned#1#2\relax{\ifx#1-\textminus\else+\fi#2}
\newcommand{\objdelta}[2]{\makebox[45pt][c]{{\scriptsize #1}\rlap{\hspace{0.2pt}\raisebox{-0.35ex}{\fontsize{6.5}{7.5}\selectfont\bfseries\boldmath$\Delta$\kern-0.3pt\objdeltasigned#2\relax}}}}

\begin{tabular*}{\textwidth}{@{\extracolsep{\fill}}ll*{9}{c}@{\kern10pt}}
\toprule
 & Method & \multicolumn{3}{c}{20\%} & \multicolumn{3}{c}{50\%} & \multicolumn{3}{c}{80\%}\\
\cmidrule(lr){3-5}\cmidrule(lr){6-8}\cmidrule(l){9-11}
& & Loc. Change & New Req. & Cancel. & Loc. Change & New Req. & Cancel. & Loc. Change & New Req. & Cancel.\\
\specialrule{0.4pt}{2pt}{0pt}
\rowcolor{gray!15}\multicolumn{11}{@{}l@{}}{\strut GLS}\\
\specialrule{0.4pt}{0pt}{2pt}
& EoH  & \objmean{8.822} & \objmean{10.090} & \objmean{7.885} & \objmean{5.619} & \objmean{7.365} & \objmean{4.965} & \objmean{2.319} & \objmean{4.235} & \objmean{2.045}\\
 & \hspace{0.5em}+\method{}  & \objdelta{8.946}{+0.124} & \objdelta{10.240}{+0.150} & \objdelta{7.979}{+0.094} & \objdelta{5.709}{+0.090} & \objdelta{7.507}{+0.142} & \objdelta{5.001}{+0.036} & \objdelta{2.338}{+0.019} & \objdelta{4.279}{+0.044} & \objdelta{2.051}{+0.006}\\
\addlinespace[0.8pt]
 & MEoH  & \objmean{9.063} & \objmean{10.248} & \objmean{7.925} & \objmean{5.685} & \objmean{7.534} & \objmean{5.016} & \objmean{2.341} & \objmean{4.155} & \objmean{2.049}\\
 & \hspace{0.5em}+\method{}  & \objdelta{9.083}{+0.020} & \objdelta{10.251}{+0.003} & \objdelta{7.933}{+0.008} & \objdelta{5.699}{+0.014} & \objdelta{7.550}{+0.016} & \objdelta{5.020}{+0.004} & \objdelta{2.341}{+0.000} & \objdelta{4.160}{+0.005} & \objdelta{2.049}{+0.000}\\
\addlinespace[0.8pt]
 & ReEvo  & \objmean{8.916} & \objmean{10.163} & \objmean{7.876} & \objmean{5.628} & \objmean{7.460} & \objmean{4.980} & \objmean{2.314} & \objmean{4.143} & \objmean{2.047}\\
 & \hspace{0.5em}+\method{}  & \objdelta{8.935}{+0.019} & \objdelta{10.165}{+0.002} & \objdelta{7.881}{+0.005} & \objdelta{5.632}{+0.004} & \objdelta{7.477}{+0.017} & \objdelta{4.983}{+0.003} & \objdelta{2.313}{-0.001} & \objdelta{4.146}{+0.003} & \objdelta{2.047}{+0.000}\\
\addlinespace[0.8pt]
 & Hercules  & \objmean{9.020} & \objmean{10.207} & \objmean{7.914} & \objmean{5.665} & \objmean{7.533} & \objmean{4.989} & \objmean{2.314} & \objmean{4.142} & \objmean{2.047}\\
\multirow{-8}{*}{\rotatebox[origin=c]{90}{TSP200}} & \hspace{0.5em}+\method{}  & \objdelta{9.025}{+0.005} & \objdelta{10.210}{+0.003} & \objdelta{7.922}{+0.008} & \objdelta{5.665}{+0.000} & \objdelta{7.529}{-0.004} & \objdelta{4.987}{-0.002} & \objdelta{2.313}{-0.001} & \objdelta{4.146}{+0.004} & \objdelta{2.047}{+0.000}\\
\midrule
 & EoH  & \objmean{13.965} & \objmean{15.798} & \objmean{12.042} & \objmean{8.840} & \objmean{11.473} & \objmean{7.571} & \objmean{3.635} & \objmean{6.408} & \objmean{3.051}\\
 & \hspace{0.5em}+\method{}  & \objdelta{14.148}{+0.183} & \objdelta{15.900}{+0.102} & \objdelta{12.207}{+0.165} & \objdelta{9.010}{+0.170} & \objdelta{11.654}{+0.181} & \objdelta{7.641}{+0.070} & \objdelta{3.695}{+0.060} & \objdelta{6.482}{+0.074} & \objdelta{3.065}{+0.014}\\
\addlinespace[0.8pt]
 & MEoH  & \objmean{14.288} & \objmean{16.026} & \objmean{12.232} & \objmean{8.912} & \objmean{11.601} & \objmean{7.624} & \objmean{3.643} & \objmean{6.370} & \objmean{3.059}\\
 & \hspace{0.5em}+\method{}  & \objdelta{14.437}{+0.149} & \objdelta{16.084}{+0.058} & \objdelta{12.321}{+0.089} & \objdelta{9.060}{+0.148} & \objdelta{11.703}{+0.102} & \objdelta{7.660}{+0.036} & \objdelta{3.678}{+0.035} & \objdelta{6.408}{+0.038} & \objdelta{3.067}{+0.008}\\
\addlinespace[0.8pt]
 & ReEvo  & \objmean{14.036} & \objmean{15.752} & \objmean{12.150} & \objmean{8.907} & \objmean{11.566} & \objmean{7.626} & \objmean{3.628} & \objmean{6.292} & \objmean{3.062}\\
 & \hspace{0.5em}+\method{}  & \objdelta{14.070}{+0.034} & \objdelta{15.766}{+0.014} & \objdelta{12.165}{+0.015} & \objdelta{8.932}{+0.025} & \objdelta{11.589}{+0.023} & \objdelta{7.629}{+0.003} & \objdelta{3.637}{+0.009} & \objdelta{6.294}{+0.002} & \objdelta{3.064}{+0.002}\\
\addlinespace[0.8pt]
 & Hercules  & \objmean{13.980} & \objmean{15.772} & \objmean{12.092} & \objmean{8.834} & \objmean{11.518} & \objmean{7.597} & \objmean{3.636} & \objmean{6.351} & \objmean{3.057}\\
\multirow{-8}{*}{\rotatebox[origin=c]{90}{TSP500}} & \hspace{0.5em}+\method{}  & \objdelta{14.154}{+0.174} & \objdelta{15.853}{+0.081} & \objdelta{12.202}{+0.110} & \objdelta{8.994}{+0.160} & \objdelta{11.634}{+0.116} & \objdelta{7.645}{+0.048} & \objdelta{3.674}{+0.038} & \objdelta{6.386}{+0.035} & \objdelta{3.064}{+0.007}\\
\specialrule{0.4pt}{2pt}{0pt}
\rowcolor{gray!15}\multicolumn{11}{@{}l@{}}{\strut ILS}\\
\specialrule{0.4pt}{0pt}{2pt}
& EoH & \objmean{8.803} & \objmean{10.096} & \objmean{7.808} & \objmean{5.519} & \objmean{7.398} & \objmean{4.901} & \objmean{2.265} & \objmean{4.052} & \objmean{1.990}\\
 & \hspace{0.5em}+\method{} & \objdelta{8.895}{+0.092} & \objdelta{10.155}{+0.059} & \objdelta{7.832}{+0.024} & \objdelta{5.535}{+0.016} & \objdelta{7.426}{+0.028} & \objdelta{4.903}{+0.002} & \objdelta{2.265}{+0.000} & \objdelta{4.054}{+0.002} & \objdelta{1.990}{+0.000}\\
\addlinespace[0.8pt]
 & MEoH & \objmean{9.104} & \objmean{10.266} & \objmean{7.913} & \objmean{5.632} & \objmean{7.507} & \objmean{4.934} & \objmean{2.280} & \objmean{4.078} & \objmean{1.993}\\
 & \hspace{0.5em}+\method{}  & \objdelta{9.206}{+0.102} & \objdelta{10.338}{+0.072} & \objdelta{7.946}{+0.033} & \objdelta{5.747}{+0.115} & \objdelta{7.555}{+0.048} & \objdelta{4.939}{+0.005} & \objdelta{2.280}{+0.000} & \objdelta{4.090}{+0.012} & \objdelta{1.993}{+0.000}\\
\addlinespace[0.8pt]
 & ReEvo & \objmean{8.874} & \objmean{10.139} & \objmean{7.829} & \objmean{5.537} & \objmean{7.421} & \objmean{4.903} & \objmean{2.266} & \objmean{4.054} & \objmean{1.990}\\
 & \hspace{0.5em}+\method{} & \objdelta{8.824}{-0.050} & \objdelta{10.114}{-0.025} & \objdelta{7.821}{-0.008} & \objdelta{5.520}{-0.017} & \objdelta{7.407}{-0.014} & \objdelta{4.904}{+0.001} & \objdelta{2.265}{-0.001} & \objdelta{4.052}{-0.002} & \objdelta{1.990}{+0.000}\\
\addlinespace[0.8pt]
 & Hercules & \objmean{8.811} & \objmean{10.092} & \objmean{7.814} & \objmean{5.516} & \objmean{7.401} & \objmean{4.901} & \objmean{2.265} & \objmean{4.052} & \objmean{1.990}\\
\multirow{-8}{*}{\rotatebox[origin=c]{90}{TSP200}} & \hspace{0.5em}+\method{} & \objdelta{8.953}{+0.142} & \objdelta{10.201}{+0.109} & \objdelta{7.855}{+0.041} & \objdelta{5.563}{+0.047} & \objdelta{7.446}{+0.045} & \objdelta{4.914}{+0.013} & \objdelta{2.266}{+0.001} & \objdelta{4.057}{+0.005} & \objdelta{1.990}{+0.000}\\
\midrule
 & EoH  & \objmean{14.103} & \objmean{15.767} & \objmean{12.114} & \objmean{8.821} & \objmean{11.496} & \objmean{7.584} & \objmean{3.641} & \objmean{6.213} & \objmean{3.068}\\
 & \hspace{0.5em}+\method{}  & \objdelta{14.259}{+0.156} & \objdelta{15.800}{+0.033} & \objdelta{12.185}{+0.071} & \objdelta{8.943}{+0.122} & \objdelta{11.555}{+0.059} & \objdelta{7.612}{+0.028} & \objdelta{3.662}{+0.021} & \objdelta{6.242}{+0.029} & \objdelta{3.073}{+0.005}\\
\addlinespace[0.8pt]
 & MEoH  & \objmean{14.466} & \objmean{16.019} & \objmean{12.336} & \objmean{9.072} & \objmean{11.717} & \objmean{7.628} & \objmean{3.761} & \objmean{6.353} & \objmean{3.069}\\
 & \hspace{0.5em}+\method{}  & \objdelta{14.497}{+0.031} & \objdelta{16.008}{-0.011} & \objdelta{12.333}{-0.003} & \objdelta{9.073}{+0.001} & \objdelta{11.696}{-0.021} & \objdelta{7.621}{-0.007} & \objdelta{3.756}{-0.005} & \objdelta{6.347}{-0.006} & \objdelta{3.069}{+0.000}\\
\addlinespace[0.8pt]
 & ReEvo & \objmean{14.158} & \objmean{15.790} & \objmean{12.165} & \objmean{8.873} & \objmean{11.537} & \objmean{7.603} & \objmean{3.644} & \objmean{6.233} & \objmean{3.069}\\
 & \hspace{0.5em}+\method{} & \objdelta{14.133}{-0.025} & \objdelta{15.757}{-0.033} & \objdelta{12.140}{-0.025} & \objdelta{8.861}{-0.012} & \objdelta{11.502}{-0.035} & \objdelta{7.586}{-0.017} & \objdelta{3.636}{-0.008} & \objdelta{6.229}{-0.004} & \objdelta{3.069}{+0.000}\\
\addlinespace[0.8pt]
 & Hercules & \objmean{14.117} & \objmean{15.785} & \objmean{12.157} & \objmean{8.894} & \objmean{11.535} & \objmean{7.601} & \objmean{3.654} & \objmean{6.234} & \objmean{3.071}\\
\multirow{-8}{*}{\rotatebox[origin=c]{90}{TSP500}} & \hspace{0.5em}+\method{} & \objdelta{14.204}{+0.087} & \objdelta{15.798}{+0.013} & \objdelta{12.173}{+0.016} & \objdelta{8.907}{+0.013} & \objdelta{11.527}{-0.008} & \objdelta{7.606}{+0.005} & \objdelta{3.652}{-0.002} & \objdelta{6.225}{-0.009} & \objdelta{3.070}{-0.001}\\
\bottomrule
\end{tabular*}
\endgroup
\end{table*}

\subsection{Main Results}
\label{sec:main-results}
\label{sec:ils-extension}
\label{sec:additional-search-operators}

\looseness=-1 Tables~\ref{tab:repo-large-scale} and~\ref{tab:objective-results} report solving times and final objective values, respectively, for four AHD methods and their \method{} variants under GLS and ILS. The evaluation covers TSP200 and TSP500 with three change types at three tour-progress levels. Under the same stopping criteria, \method{} reduces re-solving time in every tested configuration, achieving average speedups of $2.2\times$ under GLS and $4.4\times$ under ILS, with maximum relative increases in mean final tour length of only approximately 1.93\% and 2.04\%, respectively. The larger gains under ILS may reflect its direct tour perturbations, which allow runtime-aware selection to influence the subsequent local-search workload more directly than distance-guided perturbations in GLS. Some configurations run faster on TSP500 than on TSP200 because runtime depends on the local-search workload induced by the separately evolved heuristics as well as instance size.

\looseness=-1 Across change types, new requests generally yield the largest absolute time savings. New requests enlarge the remaining problem and introduce additional insertion and ordering decisions, making efficient search particularly valuable. Consistent gains under cancellations and location changes further demonstrate applicability to changes in both problem size and geometry. Across execution stages, absolute time savings are greatest at 20\% progress and decrease as the remaining problem shrinks. At 80\% progress, fewer unvisited nodes mean fewer nodes affected under the same change settings. The reduced workload limits the absolute time savings, while the smaller residual problems are consistent with nearly identical objective values in some settings. Gains also depend on the baseline's existing emphasis on efficiency. EoH achieves the largest average speedup in both frameworks, whereas MEoH already incorporates runtime into evolution and exhibits smaller gains. This suggests that the benefit of \method{} depends partly on the extent to which the underlying search and selection mechanisms already promote solving efficiency.

\begin{table*}[!tbp]
\renewcommand{\arraystretch}{0.85}
\centering
\caption{Solving-time ablation results (s) for Hercules+\method{} within GLS on TSP500 at progress ratios of 20\%, 50\%, and 80\%. For each scenario, solving times are first averaged over successful instances in each run, and the reported value is the mean of these instance averages across three runs. Parentheses report the solving-time ratio relative to Full.}
\label{tab:ablation}

\begingroup
\small
\setlength{\tabcolsep}{0pt}
\begin{tabular*}{\textwidth}{@{\extracolsep{\fill}}l*{9}{c}@{}}
\toprule
Variant & \multicolumn{3}{c}{20\%} & \multicolumn{3}{c}{50\%} & \multicolumn{3}{c}{80\%}\\
\cmidrule(lr){2-4}\cmidrule(lr){5-7}\cmidrule(l){8-10}
& Loc. Change & New Req. & Cancel. & Loc. Change & New Req. & Cancel. & Loc. Change & New Req. & Cancel.\\
\specialrule{0.4pt}{2pt}{0pt}
\rowcolor{gray!15}\multicolumn{10}{l}{\strut Hercules}\\
\specialrule{0.4pt}{0pt}{2pt}
Full & \textbf{2.045} & \textbf{3.489} & \textbf{1.345} & \textbf{0.878} & \textbf{1.342} & \textbf{0.697} & \textbf{0.188} & \textbf{0.200} & \textbf{0.090}\\
w/o Val. & \ablationratio{2.577}{1.26} & \ablationratio{4.187}{1.2} & \ablationratio{1.775}{1.32} & \ablationratio{1.089}{1.24} & \ablationratio{1.597}{1.19} & \ablationratio{0.795}{1.14} & \ablationratio{0.224}{1.19} & \ablationratio{0.232}{1.16} & \ablationratio{0.101}{1.12}\\
w/o Time$_1$ & \ablationratio{3.579}{1.75} & \ablationratio{5.234}{1.5} & \ablationratio{2.179}{1.62} & \ablationratio{1.370}{1.56} & \ablationratio{2.120}{1.58} & \ablationratio{0.934}{1.34} & \ablationratio{0.269}{1.43} & \ablationratio{0.330}{1.65} & \ablationratio{0.145}{1.61}\\
w/o Time$_2$ & \ablationratio{2.495}{1.22} & \ablationratio{4.605}{1.32} & \ablationratio{1.601}{1.19} & \ablationratio{1.027}{1.17} & \ablationratio{1.785}{1.33} & \ablationratio{0.920}{1.32} & \ablationratio{0.224}{1.19} & \ablationratio{0.236}{1.18} & \ablationratio{0.098}{1.09}\\
w/o $\sigma_t$ & \ablationratio{2.536}{1.24} & \ablationratio{4.222}{1.21} & \ablationratio{1.412}{1.05} & \ablationratio{1.045}{1.19} & \ablationratio{1.651}{1.23} & \ablationratio{0.899}{1.29} & \ablationratio{0.250}{1.33} & \ablationratio{0.262}{1.31} & \ablationratio{0.105}{1.17}\\
w/o $w_t$ & \ablationratio{2.802}{1.37} & \ablationratio{4.954}{1.42} & \ablationratio{1.762}{1.31} & \ablationratio{1.264}{1.44} & \ablationratio{2.040}{1.52} & \ablationratio{0.920}{1.32} & \ablationratio{0.227}{1.21} & \ablationratio{0.262}{1.31} & \ablationratio{0.107}{1.19}\\
\bottomrule
\end{tabular*}
\endgroup

\end{table*}

\begin{figure*}[!tbp]
\centering
\includegraphics[width=0.95\textwidth]{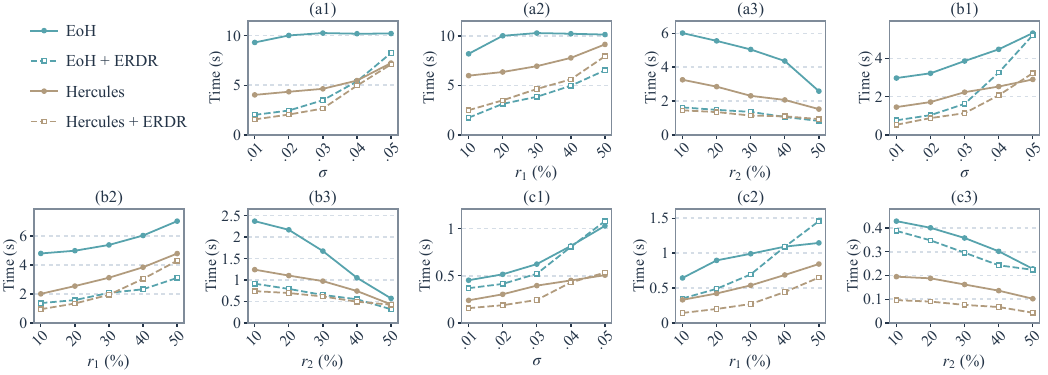}
\caption{Mean solving times of EoH, Hercules, and their \method{} variants within GLS on TSP500. Panels (a1)--(a3), (b1)--(b3), and (c1)--(c3) correspond to 20\%, 50\%, and 80\% tour progress, respectively. Within each group, panels 1, 2, and 3 vary coordinate perturbation $\sigma$, node-addition ratio $r_1$, and node-removal ratio $r_2$ within their respective dynamic scenarios.}
\label{fig:generalization-boundary}
\end{figure*}

\subsection{Ablation Study}
\label{sec:ablation}

\looseness=-1 We compare the complete \method{} with five ablation variants for Hercules within GLS. w/o Val. removes validation-guided reselection while retaining the two time criteria, whereas w/o Time$_1$ retains reselection but removes both criteria. w/o Time$_2$ replaces inter-group variability and tail risk with mean per-instance solving time while preserving the original objectives and validation-guided reselection. w/o $\sigma_t$ and w/o $w_t$ further isolate the contributions of inter-group variability and tail risk, respectively. All variants use the same generation budget and search settings within each baseline. Table~\ref{tab:ablation} reports mean solving times. Among the evaluated Hercules ablations, the complete framework consistently achieves the lowest solving time. Removing both time criteria (w/o Time$_1$) causes the largest overall increase, indicating their substantial contribution to reducing search time. Replacing the two robustness criteria with mean solving time improves efficiency over w/o Time$_1$ but remains slower than Full across all scenarios. Removing tail risk leads to a larger overall increase than removing inter-group variability, suggesting a greater contribution from the tail-risk criterion under the evaluated settings. Removing validation-guided reselection also increases solving time, showing its additional benefit alongside the time criteria. Larger absolute time savings are observed under new requests and at early tour progress, where more nodes remain to be optimized.

\subsection{Boundary Analysis}
\label{sec:generalization-boundary}

\looseness=-1 We examine how the solving-time advantage of \method{} varies with tour progress and dynamic-change strength. The evaluation uses EoH and Hercules within GLS on TSP500. Each comparison uses the same residual instances and warm-start routes, and reports the mean solving time of each method. At each tour-progress level $\rho\in\{20\%,50\%,80\%\}$, we vary change strength within each dynamic scenario separately. We vary $\sigma$ over $\{0.01,0.02,0.03,0.04,0.05\}$ for location changes, $r_1$ over $\{10\%,20\%,30\%,40\%,50\%\}$ for new requests, and $r_2$ over the same ratios for request cancellations, keeping all other experimental settings fixed at their default values.

\looseness=-1 Figure~\ref{fig:generalization-boundary} shows that \method{} retains its solving-time advantage across most tested change strengths and progress levels, supporting the generalization of its search efficiency beyond the default scenarios. The EoH plateaus near 10 seconds in (a1) and (a2) reflect exhaustion of the search-time budget, with slight overruns caused by completing the ongoing iteration. At high coordinate-perturbation strengths or node-addition ratios, the runtime advantage can disappear or reverse, as observed in (b1), (c1), and (c2). Large coordinate shifts or node insertions may reorder candidate edges and trigger cascading tour adjustments. If \method{} perturbs more connections, these cascades can incur greater local-search costs than more conservative baseline perturbations, reversing its runtime advantage. Under extensive cancellations and late tour progress, the smaller residual task reduces the search workload and the absolute room for improvement, producing short runtimes and small differences in several settings.

\begin{figure}[!htbp]
\centering
\begin{minipage}{\columnwidth}
\centering
\includegraphics[width=0.95\linewidth]{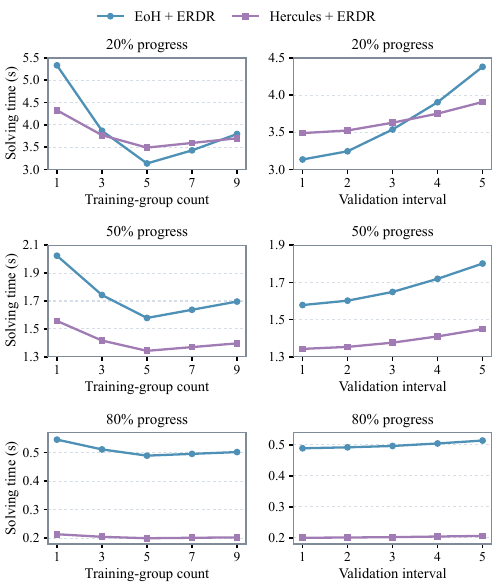}
\caption{Solving-time sensitivity to the training-group count (left) and validation interval (right) within GLS on TSP500 with new requests added at a node-addition ratio of 20\%.  The validation interval counts population updates between consecutive validation-guided reselections. Rows correspond to 20\%, 50\%, and 80\% tour progress. Panels use separate vertical scales. Blue and purple denote EoH+\method{} and Hercules+\method{}, respectively. Lower values are better.}
\label{fig:supp-parameter-analysis}
\end{minipage}
\end{figure}

\subsection{Parameter Analysis}
\label{app:parameter-sensitivity}

We examine the sensitivity of \method{} to the training--validation allocation and validation interval by varying the training-group count and  the number of population updates between consecutive validation-guided reselections. We vary one parameter at a time within GLS on TSP500 with new requests added at a node-addition ratio of 20\%, evaluating EoH+\method{} and Hercules+\method{} at 20\%, 50\%, and 80\% tour progress.

\looseness=-1 Figure~\ref{fig:supp-parameter-analysis} shows that solving time initially decreases and then increases as the training-group count increases. Increasing the training allocation expands the instance coverage of routine candidate evaluation while leaving fewer instances for validation-guided reselection. An equal allocation between training and validation yields the lowest solving time for both methods across the three progress levels.  Validation at every population update or every two updates yields similar solving times. Longer intervals increase solving time, with the largest absolute increases at 20\% tour progress, indicating that frequent validation-guided reselection benefits re-solving efficiency in this setting.

\begin{table}[!htbp]
\renewcommand{\arraystretch}{0.85}
\centering
\caption{Search time (s) and iteration counts to first match or improve on the higher final objective value in each baseline--\method{} pair across dynamic scenarios at 20\% tour progress. Entries are means over paired successful instances. Lower values within each baseline pair are shown in bold.}
\label{tab:common-objective}
\label{tab:common-objective-ils}
\begingroup
\footnotesize
\setlength{\tabcolsep}{2pt}
\begin{tabular*}{\columnwidth}{@{\extracolsep{\fill}}l*{6}{c}@{}}
\toprule
Method & \multicolumn{2}{c}{Loc. change} & \multicolumn{2}{c}{New Req.} & \multicolumn{2}{c}{Cancel.}\\
\cmidrule(lr){2-3}\cmidrule(lr){4-5}\cmidrule(l){6-7}
 & Time (s) $\downarrow$ & Iter. $\downarrow$ & Time (s) $\downarrow$ & Iter. $\downarrow$ & Time (s) $\downarrow$ & Iter. $\downarrow$\\
\specialrule{0.4pt}{2pt}{0pt}
\rowcolor{gray!15}\multicolumn{7}{l}{\strut GLS -- TSP200}\\
\specialrule{0.4pt}{0pt}{2pt}
EoH & 0.3592 & 40.42 & 0.2818 & 37.00 & 0.0968 & 18.14\\
+\method{} & \textbf{0.0158} & \textbf{23.44} & \textbf{0.0144} & \textbf{21.84} & \textbf{0.0132} & \textbf{5.30}\\
ReEvo & 0.1820 & 242.68 & 0.2119 & 196.84 & 0.0903 & 195.00\\
+\method{} & \textbf{0.0780} & \textbf{203.38} & \textbf{0.0973} & \textbf{150.50} & \textbf{0.0420} & \textbf{167.98}\\
\specialrule{0.4pt}{2pt}{0pt}
\rowcolor{gray!15}\multicolumn{7}{l}{\strut GLS -- TSP500}\\
\specialrule{0.4pt}{0pt}{2pt}
EoH & 0.5057 & 93.38 & 2.1124 & 109.64 & 0.5624 & 88.10\\
+\method{} & \textbf{0.1680} & \textbf{16.58} & \textbf{0.6982} & \textbf{93.04} & \textbf{0.2834} & \textbf{50.02}\\
ReEvo & 0.6162 & 397.80 & 0.7055 & 249.76 & 0.3011 & 235.94\\
+\method{} & \textbf{0.5418} & \textbf{312.04} & \textbf{0.4763} & \textbf{230.88} & \textbf{0.1954} & \textbf{220.40}\\
\specialrule{0.4pt}{2pt}{0pt}
\rowcolor{gray!15}\multicolumn{7}{l}{\strut ILS -- TSP200}\\
\specialrule{0.4pt}{0pt}{2pt}
EoH & 0.5321 & 223.46 & 1.0484 & 339.80 & 0.6390 & 270.16\\
+\method{} & \textbf{0.2085} & \textbf{87.98} & \textbf{0.3394} & \textbf{125.58} & \textbf{0.1567} & \textbf{131.70}\\
ReEvo & 4.8729 & 374.52 & 5.2356 & 305.42 & 2.1675 & 195.32\\
+\method{} & \textbf{0.4074} & \textbf{229.24} & \textbf{0.4167} & \textbf{174.08} & \textbf{0.1668} & \textbf{134.96}\\
\specialrule{0.4pt}{2pt}{0pt}
\rowcolor{gray!15}\multicolumn{7}{l}{\strut ILS -- TSP500}\\
\specialrule{0.4pt}{0pt}{2pt}
EoH & 1.1277 & 201.58 & 2.4566 & 285.80 & 1.2149 & 198.16\\
+\method{} & \textbf{0.7821} & \textbf{87.98} & \textbf{0.5478} & \textbf{144.32} & \textbf{0.5897} & \textbf{118.70}\\
ReEvo & 1.4465 & 415.38 & 1.6231 & 341.48 & 0.6422 & 337.14\\
+\method{} & \textbf{1.3790} & \textbf{386.26} & \textbf{1.6108} & \textbf{284.30} & \textbf{0.6224} & \textbf{271.36}\\
\bottomrule
\end{tabular*}
\endgroup
\end{table}

\subsection{Search Efficiency Analysis}
\label{sec:common-objective}
\looseness=-1 We analyze search trajectories to examine whether \method{} indeed accelerates the search process. We compare each baseline and its \method{} variant in the time and iterations needed to first match or improve on the higher of their final objective values on each instance within the search budget. Following the default experimental settings, we evaluate EoH and ReEvo under GLS and ILS on TSP200 and TSP500 at 20\% tour progress. Table~\ref{tab:common-objective} reports reductions in both measures across the evaluated configurations. These results show that \method{} reaches the paired common objectives in less time and with fewer search iterations. Across dynamic events, new requests generally yield greater absolute time savings than cancellations. This may be due to the additional insertion and ordering decisions introduced by new requests, which leave more costly exploration for \method{} to reduce, whereas cancellations shrink the remaining task and limit the available savings. Search progress also differs across methods. EoH generally exhibits larger relative iteration reductions, suggesting that its evolved heuristics may contain more redundant exploration that runtime-aware selection can remove. ReEvo retains higher iteration counts in most settings, which may reflect improvements spread across more search steps and leave less room to shorten the trajectory.

\begin{figure*}[!tbp]
\centering
\includegraphics[width=0.97\textwidth]{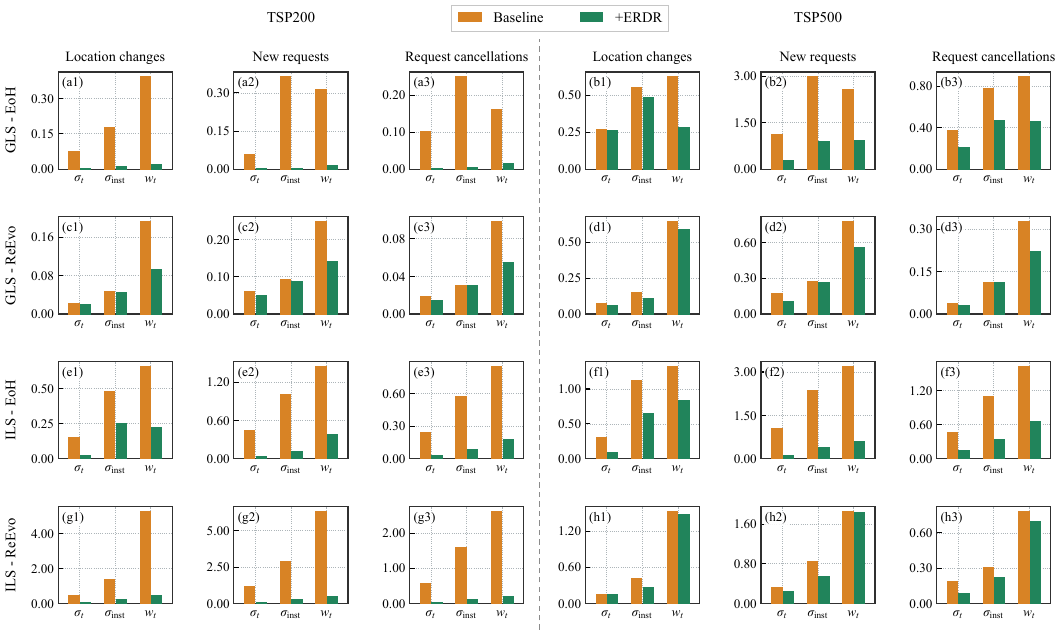}
\caption{Variability and tail costs of time to reach the paired common objectives after dynamic changes at 20\% tour progress. Rows separate search frameworks and baselines. The left and right groups report TSP200 and TSP500, each with columns for location changes, new requests, and request cancellations. Bars show $\sigma_t$, $\sigma_{\mathrm{inst}}$, and $w_t$ for the baseline (orange) and its \method{} variant (green). All values are in seconds.}
\label{fig:runtime-stability}
\label{fig:runtime-stability-progress}
\label{fig:runtime-stability-progress-500}
\label{tab:runtime-stability}
\label{tab:runtime-stability-ils}
\end{figure*}

\subsection{Runtime Stability and Tail Performance}
\label{sec:runtime-stability}
\looseness=-1 We further examine whether the acceleration achieved by \method{} is accompanied by reduced variation in search time across instances and lower costs in slower instance groups during dynamic re-solving. Following the default experimental settings, we compare EoH and ReEvo with their \method{} variants under GLS and ILS on TSP200 and TSP500 at 20\% tour progress. Using the time required by each solver to first match or improve on the paired common objective defined in Section~\ref{sec:common-objective}, we compute inter-group standard deviation $\sigma_t$ and tail mean $w_t$ to characterize runtime variability and slow-group costs, respectively. We also report instance-level standard deviation $\sigma_{\mathrm{inst}}$. Results use one evaluation run per configuration.

\looseness=-1 Figure~\ref{fig:runtime-stability} shows that new requests generally incur greater inter-group variability and tail costs, consistent with the increased search burden from additional nodes. \method{} reduces both metrics in this scenario, mitigating uneven runtimes and high slow-group costs as the remaining task expands. Request cancellations generally have lower tail costs than those observed with new requests, and \method{} further reduces runtimes in the slower groups. Under location changes, both metrics also improve with unchanged node counts, extending the gains to re-solving after spatial changes. These results indicate that \method{} alleviates uneven runtimes and high costs in slower instance groups across dynamic scenarios, contributing to more efficient dynamic re-solving.

\begin{table}[!htbp]
\renewcommand{\arraystretch}{0.85}
\centering
\caption{Mean end-to-end evolution time (h) over three runs on TSP200 and TSP500 under GLS and ILS. Each scale compares the baseline with baseline+\method{}.}
\label{tab:evolution-cost}
\begingroup
\footnotesize
\setlength{\tabcolsep}{1pt}
\begin{tabular*}{\columnwidth}{@{\extracolsep{\fill}}ll*{8}{r}@{}}
\toprule
 & & \multicolumn{2}{c}{EoH} & \multicolumn{2}{c}{Hercules} & \multicolumn{2}{c}{ReEvo} & \multicolumn{2}{c}{MEoH}\\
\cmidrule(lr){3-4}\cmidrule(lr){5-6}\cmidrule(lr){7-8}\cmidrule(l){9-10}
Scale & Variant & GLS & ILS & GLS & ILS & GLS & ILS & GLS & ILS\\
\midrule
\multirow{2}{*}{TSP200} & Baseline & 2.60 & 12.49 & 0.25 & 1.03 & 0.39 & 2.61 & 2.80 & 11.26\\
 & +\method{} & 2.40 & 11.09 & 0.23 & 1.24 & 0.33 & 2.01 & 2.62 & 11.35\\
\midrule
\multirow{2}{*}{TSP500} & Baseline & 4.30 & 22.06 & 0.55 & 7.34 & 0.84 & 7.76 & 4.65 & 19.82\\
 & +\method{} & 3.93 & 19.95 & 0.50 & 5.91 & 0.75 & 6.80 & 4.25 & 18.68\\
\bottomrule
\end{tabular*}
\endgroup
\end{table}

\begin{table*}[!tbp]
\renewcommand{\arraystretch}{0.85}
\centering
\caption{Total search time (s) of Hercules, ReEvo, and their \method{} variants on six TSPLIB instances within GLS. The heuristics are evolved on TSP200 instances and directly evaluated on dynamic tasks derived from each TSPLIB instance using the default search settings. Within each baseline pair, results with lower times are shown in bold.}
\label{tab:tsplib-selected}
\begingroup
\footnotesize
\setlength{\tabcolsep}{0pt}
\begin{tabular*}{0.95\textwidth}{@{\extracolsep{\fill}}p{0.085\textwidth}p{0.105\textwidth}*{9}{c}@{}}
\toprule
Instance & Method & \multicolumn{3}{c}{20\%} & \multicolumn{3}{c}{50\%} & \multicolumn{3}{c}{80\%}\\
\cmidrule(lr){3-5}\cmidrule(lr){6-8}\cmidrule(l){9-11}
 & & Loc. Change & New Req. & Cancel. & Loc. Change & New Req. & Cancel. & Loc. Change & New Req. & Cancel.\\
\midrule
\multirow{4}{*}{\shortstack[l]{st70\\($n=70$)}} & Hercules & 0.078 & 0.104 & 0.053 & 0.036 & 0.057 & 0.029 & 0.014 & 0.016 & 0.012\\
 & \hspace{0.5em}+\method{} & \textbf{0.047} & \textbf{0.057} & \textbf{0.030} & \textbf{0.024} & \textbf{0.029} & \textbf{0.016} & \textbf{0.012} & \textbf{0.011} & \textbf{0.009}\\
\addlinespace[0.8pt]
 & ReEvo & 0.061 & 0.095 & 0.043 & 0.028 & 0.041 & 0.021 & 0.012 & \textbf{0.013} & 0.009\\
 & \hspace{0.5em}+\method{} & \textbf{0.042} & \textbf{0.063} & \textbf{0.027} & \textbf{0.021} & \textbf{0.026} & \textbf{0.017} & \textbf{0.009} & \textbf{0.013} & \textbf{0.007}\\
\midrule
\multirow{4}{*}{\shortstack[l]{pr76\\($n=76$)}} & Hercules & 0.062 & 0.075 & 0.045 & 0.033 & 0.042 & 0.026 & 0.014 & 0.015 & 0.013\\
 & \hspace{0.5em}+\method{} & \textbf{0.044} & \textbf{0.057} & \textbf{0.032} & \textbf{0.023} & \textbf{0.029} & \textbf{0.017} & \textbf{0.011} & \textbf{0.011} & \textbf{0.008}\\
\addlinespace[0.8pt]
 & ReEvo & 0.070 & 0.063 & 0.042 & 0.038 & 0.048 & \textbf{0.020} & 0.011 & 0.011 & 0.009\\
 & \hspace{0.5em}+\method{}  & \textbf{0.044} & \textbf{0.057} & \textbf{0.032} & \textbf{0.023} & \textbf{0.030} & 0.024 & \textbf{0.009} & \textbf{0.010} & \textbf{0.008}\\
\midrule
\multirow{4}{*}{\shortstack[l]{ch150\\($n=150$)}} & Hercules & 0.592 & 0.854 & 0.382 & 0.258 & 0.344 & 0.172 & 0.053 & 0.066 & 0.035\\
 & \hspace{0.5em}+\method{} & \textbf{0.135} & \textbf{0.185} & \textbf{0.085} & \textbf{0.058} & \textbf{0.082} & \textbf{0.041} & \textbf{0.016} & \textbf{0.021} & \textbf{0.013}\\
\addlinespace[0.8pt]
 & ReEvo & 0.603 & 0.805 & 0.441 & 0.278 & 0.320 & 0.182 & 0.050 & 0.070 & 0.034\\
 & \hspace{0.5em}+\method{}& \textbf{0.150} & \textbf{0.240} & \textbf{0.115} & \textbf{0.133} & \textbf{0.093} & \textbf{0.048} & \textbf{0.026} & \textbf{0.023} & \textbf{0.016}
\\
\midrule
\multirow{4}{*}{\shortstack[l]{a280\\($n=280$)}} & Hercules & 0.681 & 0.778 & 0.346 & 0.285 & 0.363 & 0.151 & 0.056 & 0.072 & 0.043\\
 & \hspace{0.5em}+\method{} & \textbf{0.475} & \textbf{0.650} & \textbf{0.312} & \textbf{0.175} & \textbf{0.281} & \textbf{0.122} & \textbf{0.040} & \textbf{0.061} & \textbf{0.029}\\
\addlinespace[0.8pt]
 & ReEvo & 0.639 & 0.727 & 0.389 & 0.332 & 0.352 & 0.152 & 0.048 & 0.070 & 0.043\\
 & \hspace{0.5em}+\method{} & \textbf{0.501} & \textbf{0.621} & \textbf{0.284} & \textbf{0.190} & \textbf{0.276} & \textbf{0.120} & \textbf{0.040} & \textbf{0.055} & \textbf{0.029}\\
\midrule
\multirow{4}{*}{\shortstack[l]{rd400\\($n=400$)}} & Hercules & 1.157 & 1.748 & 0.702 & 0.420 & 0.715 & 0.301 & 0.087 & 0.109 & 0.066\\
 & \hspace{0.5em}+\method{} & \textbf{1.030} & \textbf{1.499} & \textbf{0.635} & \textbf{0.387} & \textbf{0.557} & \textbf{0.245} & \textbf{0.075} & \textbf{0.107} & \textbf{0.049}\\
\addlinespace[0.8pt]
 & ReEvo & 1.393 & 2.150 & 0.887 & 0.579 & 0.689 & 0.351 & 0.123 & 0.150 & 0.075\\
 & \hspace{0.5em}+\method{} & \textbf{1.102} & \textbf{1.584} & \textbf{0.687} & \textbf{0.422} & \textbf{0.611} & \textbf{0.282} & \textbf{0.079} & \textbf{0.104} & \textbf{0.060}\\
\midrule
\multirow{4}{*}{\shortstack[l]{d493\\($n=493$)}} & Hercules & 5.095 & 6.995 & 3.141 & 1.885 & 2.688 & 1.092 & 0.322 & 0.484 & 0.190\\
 & \hspace{0.5em}+\method{} & \textbf{1.797} & \textbf{2.073} & \textbf{1.012} & \textbf{0.767} & \textbf{0.873} & \textbf{0.343} & \textbf{0.112} & \textbf{0.175} & \textbf{0.074}\\
\addlinespace[0.8pt]
 & ReEvo & 2.845 & 3.071 & 2.973 & 1.619 & 1.817 & 0.415 & 0.218 & 0.265 & 0.104\\
 & \hspace{0.5em}+\method{} & \textbf{1.680} & \textbf{2.036} & \textbf{0.964} & \textbf{0.649} & \textbf{0.785} & \textbf{0.397} & \textbf{0.114} & \textbf{0.152} & \textbf{0.070}\\
\bottomrule
\end{tabular*}
\endgroup
\end{table*}

\begin{figure*}[!tbp]
\centering
\includegraphics[width=0.92\textwidth,trim=0 15bp 0 3bp,clip]{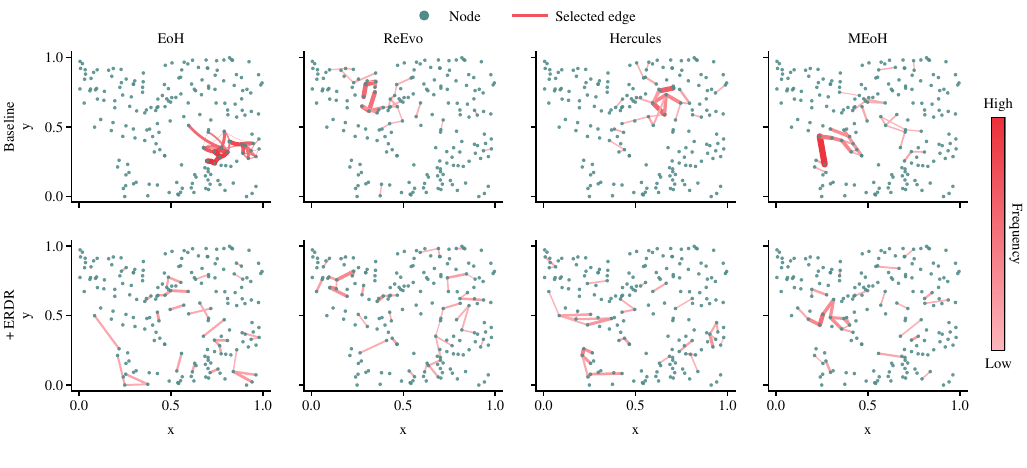}
\caption{Penalty-edge selection coverage within GLS on TSP200 with new requests added at 20\% tour progress using a node-addition ratio of 20\%. Columns show EoH, ReEvo, Hercules, and MEoH, with baselines above and their \method{} variants below. For a common spatial reference, node coordinates are aligned to the EoH baseline and edge endpoints are mapped to this layout, preserving edge colors and widths.}
\label{fig:supp-search-coverage}
\end{figure*}

\begin{figure}[!htbp]
\centering
\begingroup
\begin{minipage}{0.97\columnwidth}
\noindent\colorbox{orange!15}{\parbox{\dimexpr\linewidth-2\fboxsep\relax}{\centering\sffamily EoH}}
\begin{lstlisting}[style=staticcode,basicstyle=\ttfamily\fontsize{7.5}{8.5}\selectfont]
q_low = np.quantile(edge_n_used, 0.33)
q_high = np.quantile(edge_n_used, 0.66)
# <-- thresholds from sparse N x N matrix
for i in range(len(local_opt_tour)):
    u = local_opt_tour[i]
    v = local_opt_tour[(i + 1) %
                       len(local_opt_tour)]
    usage_count = edge_n_used[u, v]
    if usage_count < q_low:
        updated_edge_distance[u, v] *= 0.75
    elif usage_count > q_high:
        updated_edge_distance[u, v] *= 1.10
    updated_edge_distance[u, v] *= 0.97
    # <-- fixed multipliers after global calibration
\end{lstlisting}
\par
\noindent\colorbox{green!15}{\parbox{\dimexpr\linewidth-2\fboxsep\relax}{\centering\sffamily EoH+\method{}}}
\begin{lstlisting}[style=erdrcode,basicstyle=\ttfamily\fontsize{7.5}{8.5}\selectfont]
updated_edge_distance = edge_distance.copy()
total_edges = len(local_opt_tour)
average_distance = total_distance / total_edges
# <-- scale of the current tour

for i in range(total_edges):
    u = local_opt_tour[i]
    v = local_opt_tour[(i + 1) % total_edges]
    if edge_n_used[u, v] > 0:
        adjustment = (
            1 + 0.2 * np.random.uniform(-1, 1)
        ) * (
            average_distance
            / max(
                updated_edge_distance[u, v],
                1e-12
            )
        ) ** 0.5
        # <-- relative edge-to-tour calibration
        updated_edge_distance[u, v] *= adjustment

\end{lstlisting}
\end{minipage}
\endgroup
\caption{Edge-distance update logic in the final EoH (orange) and EoH+\method{} (green) solvers. Colored arrow comments highlight usage thresholds and edge-to-tour distance calibration. Only the statements needed for the comparison are shown, and local variable names are shortened.}
\label{fig:large-final-solver-code}
\end{figure}

\subsection{Evolution Overhead}
\label{app:evolution-cost}

\looseness=-1 We measure end-to-end evolution time on TSP200 and TSP500 under the matched settings and generation budgets described above, using the same 50 source instances within each comparison. Timing covers the full evolution run, including initial candidate generation, LLM calls, source-side evaluation, and survivor selection. Table~\ref{tab:evolution-cost} reports three-run means for GLS and ILS. Although ERDR introduces validation-guided reselection and runtime-robust evaluation, it generally achieves lower end-to-end evolution time across GLS and ILS. This reduction is supported by a smaller routine evaluation workload. Baselines evaluate newly generated candidates on the full source set, whereas ERDR uses only the training subset and evaluates the validation subset at scheduled population reselection steps, allowing routine evaluation savings to offset the additional validation cost. End-to-end savings also depend on the evaluation costs of intermediate candidates. API latency, retries, and evaluation timeouts can further affect this balance. For example, Hercules and MEoH incur longer evolution times under ILS on TSP200 despite faster search by their final heuristics.

\subsection{Generalization to TSPLIB Instances}
\label{sec:tsplib-generalization}

\looseness=-1 We further evaluate \method{} on dynamic tasks derived from TSPLIB, including instances from real-world applications. We evolve heuristics with Hercules, ReEvo, and their \method{} variants on TSP200 instances, then directly evaluate them on dynamic tasks derived from TSPLIB. For TSPLIB, the dynamic changes in Table~\ref{tab:dynamic-construction} are applied using each original instance's coordinate scale and bounds. Table~\ref{tab:tsplib-selected} reports results for Hercules and ReEvo within GLS on six selected instances. Across Hercules and ReEvo, \method{} achieves a mean speedup of $1.89\times$. Gains are generally larger at earlier tour progress, while small residual tasks leave less room for acceleration. The gains also vary across test instances without increasing monotonically with instance size, suggesting that spatial structure and the resulting search behavior may influence the benefit of runtime-aware heuristic selection. These results demonstrate that \method{} extends effectively to real-world application data.

\subsection{Case Study}
\label{sec:case-study}
\label{sec:search-behavior}
\label{app:search-coverage}
\paragraph{Spatial selection patterns.}

\looseness=-1 We use a representative TSP200 instance as an example to illustrate differences in penalty-edge selection between the baselines and their \method{}-enhanced variants within GLS. At 20\% tour progress, we add new requests equal in number to 20\% of the unvisited nodes, and we compare the subsequent re-solving behavior of the four baselines and their \method{}-enhanced variants. We record the edges selected for penalization by each evolved heuristic and aggregate their selection frequencies over the search. Figure~\ref{fig:supp-search-coverage} presents these patterns in a common coordinate layout, with nodes aligned to the EoH baseline and edge endpoints mapped accordingly while retaining the original stroke styles. The baseline heuristics repeatedly select penalty edges within limited regions, although the degree of concentration varies across backbones. In all four comparisons, the \method{}-enhanced heuristics cover broader regions and reduce the dominance of local high-frequency clusters while retaining backbone-specific selection patterns. These observations suggest less spatially concentrated penalty-edge selection during re-solving on this instance.

\paragraph{Evolved heuristic rules.}
\label{app:solver-behavior}

\looseness=-1 We select representative final heuristics evolved by EoH and EoH+\method{} on TSP500 as examples to examine their edge-distance update rules. Figure~\ref{fig:large-final-solver-code} presents the edge-distance update logic embedded in these final solvers, where edge costs are revised using the current locally optimal tour and accumulated edge-use counts before the next search step. EoH computes \texttt{q\_low} and \texttt{q\_high} from the full sparse $N\!\times\!N$ usage matrix before applying the multipliers \texttt{0.75}, \texttt{1.10}, and \texttt{0.97}. A sufficiently large fraction of zero entries makes both thresholds zero, so all edges with positive counts receive the same combined multiplier of $1.10\times0.97=1.067$, regardless of their count magnitudes. EoH+\method{} also selects edges with positive counts, then multiplies their costs by a random factor calibrated by the ratio of \texttt{average\_distance} to the current edge distance. The adjustment therefore varies with each edge's distance relative to the current tour. Such distance-dependent updates may help adapt search guidance after instance changes, offering a possible explanation for the observed time reductions.

\subsection{More Details and Analyses}
Due to space limitations, additional experiments and implementation details are provided in the supplementary material, available together with the code and data at \url{https://github.com/myz12138/ERDR}.

\section{Conclusion}

In this paper, we investigate inefficient re-solving by AHD-evolved TSP solvers when instances change during route execution. Our analysis suggests that ``static-overfitting'' contributes to this inefficiency, showing that evolved solvers repeatedly direct search toward overlapping local neighborhoods after such changes. 
We further link this behavior to evolutionary processes that retain heuristics whose quality gains concentrate on a few instances and are coupled with longer solving times. We propose \method{}, a plug-and-play framework combining validation-guided reselection and runtime robustness-augmented evaluation. Experiments across four AHD methods, two search frameworks (GLS and ILS), multiple TSP scales, and three change types demonstrate that \method{} reduces tour re-solving time.

% END PRESERVED MAIN CONTENT

% Start the AI declaration and references after the main paper.
% Complete the final text column using its normal stretchable spacing.
\par\pagebreak
\clearpage

\section*{AI-Generated Content Acknowledgement}
The solver evolution process in Sections~\ref{sec:methodology} and~\ref{sec:experiments} uses the gpt-6-luna API to generate and revise candidate heuristic code. ChatGPT and Codex assisted with drafting and revising text throughout the manuscript, including language clarity, paragraph structure, and terminology consistency. They also assisted with LaTeX formatting, figure and table layout throughout the manuscript, and code editing and debugging for the experiments in Section~\ref{sec:experiments}.

\bibliographystyle{IEEEtran}
\bibliography{references}

\end{document}